\documentclass[preprint,3p]{elsarticle}
\usepackage{amssymb}
\usepackage{amsmath}
\usepackage{amsfonts}
\usepackage{algorithm}
\usepackage{algorithmic}
\usepackage{float}
\usepackage{graphicx}
\usepackage{booktabs} 
\usepackage{subcaption}
\usepackage{xcolor}
\usepackage{hyperref}
\usepackage{multirow}

\begin{document} 

\begin{frontmatter}

\title{Virtual Smart Metering in District Heating Networks via Heterogeneous Spatial-Temporal Graph Neural Networks}

\author[inst1]{Keivan Faghih Niresi}
\ead{keivan.faghihniresi@epfl.ch}

\author[inst2,inst3]{Christian Møller Jensen}
\ead{cmje@es.aau.dk}

\author[inst2,inst3]{Carsten Skovmose Kallesøe}
\ead{csk@es.aau.dk}

\author[inst2]{Rafael Wisniewski}
\ead{raf@es.aau.dk}

\author[inst1]{Olga Fink\corref{cor1}}
\ead{olga.fink@epfl.ch}

\cortext[cor1]{Corresponding author}

\address[inst1]{Intelligent Maintenance and Operations Systems Lab., EPFL, Lausanne, Switzerland}
\address[inst2]{Department of Electronic Systems, Aalborg University, Aalborg, Denmark}
\address[inst3]{Grundfos Holding A/S, Denmark}

\begin{abstract}

Intelligent operation of thermal energy networks aims to improve energy efficiency, reliability, and operational flexibility through data-driven control, predictive optimization, and early fault detection. Achieving these goals relies on sufficient observability, requiring continuous and well-distributed monitoring of thermal and hydraulic states. However, district heating systems are typically sparsely instrumented and frequently affected by sensor faults, limiting monitoring. Virtual sensing offers a cost-effective means to enhance observability, yet its development and validation remain limited in practice.  Existing data-driven methods generally assume dense synchronized data, while analytical models rely on simplified hydraulic and thermal assumptions that  may not adequately capture the behavior of heterogeneous network topologies. Consequently, modeling the coupled nonlinear dependencies between pressure, flow, and temperature under realistic operating conditions remains challenging. In addition, the lack of publicly available benchmark datasets hinders systematic comparison of virtual sensing approaches. To address these challenges, we propose a heterogeneous spatial-temporal graph neural network (HSTGNN)  for constructing virtual smart heat meters. The model incorporates the functional relationships inherent in district heating networks and employs dedicated branches to learn graph structures and temporal dynamics for flow, temperature, and pressure measurements, thereby enabling the joint modeling of cross-variable and spatial correlations. To support further research, we introduce a controlled laboratory dataset collected at the Aalborg Smart Water Infrastructure Laboratory, providing synchronized high-resolution measurements representative of real operating conditions.  Extensive experiments demonstrate that the proposed approach significantly outperforms existing baselines, demonstrating the potential of graph-based learning to support efficient, reliable, and flexible operation of modern district heating systems.
\end{abstract}

\begin{keyword}
Virtual sensing, Spatial-temporal graph neural networks, District heating, Smart meters, Intelligent energy networks.
\end{keyword}

\end{frontmatter}

\section{Introduction}


The transition toward intelligent thermal energy networks aims to improve efficiency, reliability, and flexibility through predictive control, optimization, and integration of renewable heat sources. District heating systems play a central role in this transition \cite{connolly2014heat} by enabling the large-scale use of low-carbon energy streams such as geothermal energy, industrial waste heat, solar thermal installations, large heat pumps, and power-to-heat technologies \cite{lund2014fourth, werner2017district}. As decarbonization efforts intensify, modern district heating networks are becoming increasingly dynamic and heterogeneous. Fourth- and fifth-generation concepts emphasize low-temperature operation, distributed production, and prosumer participation, requiring advanced monitoring and digital control strategies to maintain stable and efficient operation \cite{hast2018district}.

Achieving such intelligent operation depends on sufficient observability of thermal and hydraulic states. Smart heat meters and supervisory sensors measure flow rate, inlet and outlet temperatures, and accumulated heat energy, supporting tasks such as billing, load estimation, fault detection, leakage identification, and pump control optimization \cite{kadlec2009data, saloux2023towards}. Pressure sensors complement these measurements by capturing hydraulic conditions and enabling detection of abnormal pressure drops and flow variations \cite{yoon2020virtual}. In practice, however, district heating systems are sparsely instrumented, and measurements are often asynchronous, noisy, or faulty. This asynchronicity is partly a consequence of how smart meters operate: because many are battery-powered, they typically transmit data infrequently and in batches, since data transmission is a major drain on battery life. This limits the ability to reliably estimate system states and constrains the deployment of advanced control and optimization strategies.


Virtual sensing has therefore emerged as a promising approach to enhance observability without extensive hardware deployment \cite{sun2021survey, fink2025physics, choi2023situ}. Existing approaches can be broadly categorized into analytical, state-estimation, and data-driven methods. Analytical thermo-hydraulic models rely on physical equations describing pressure drops, flow balance, and heat transport, and can provide interpretable estimates when detailed network parameters are available. However, their accuracy depends strongly on pipe parameters, valve states, demand profiles, and heat-loss assumptions, which are often uncertain in operational networks. State-estimation methods, such as Kalman-filtering-based approaches \cite{petersen2008kalman}, combine model predictions with measurements and can be extended to nonlinear systems through variants such as the extended or unscented Kalman filter \cite{wan2000unscented, julier1997new}. However, these methods still require a suitable state-space representation, including transition dynamics, measurement functions, noise assumptions, and, in many cases, thermo-hydraulic parameters or equivalent model structures. Constructing and calibrating such models can be challenging in district heating systems because of sparse and heterogeneous sensing, uncertain boundary conditions, changing operating regimes, and incomplete knowledge of component-level dynamics. Data-driven approaches reduce the need for explicitly specified physical models, but many treat sensor measurements as independent or homogeneous time series, limiting their ability to capture structured dependencies between thermal, hydraulic, and flow-related variables. These limitations motivate methods that can learn heterogeneous sensor interactions and complex spatial-temporal dependencies directly from operational data.

\begin{figure} \centering \includegraphics[width=0.6\linewidth]{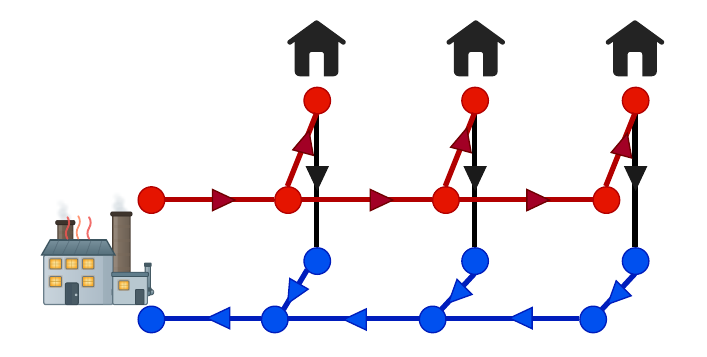} \caption{Schematic layout of the district heating network under study. The tree-structured graph illustrates the symmetrical relationship between the supply and return networks, where edge directions are inverted to reflect flow orientation.} \label{fig:tree} \end{figure}

While machine learning has shown promise for virtual sensing in district heating systems, two major challenges remain. First, the development and benchmarking of these methods are  hindered by the scarcity of high-quality, synchronized, and publicly available experimental and real-world datasets. Most research studies rely on idealized simulations that fail to capture sensor noise, measurement delays, and transient hydraulic behavior in real networks \cite{niresi2024informed}. This lack of realistic benchmarks limits reproducibility and makes it difficult to assess how models perform under practical operating conditions. Second, many data-driven models approaches treat sensors as independent time series, neglecting the functional relationships imposed by the pipe network \cite{li2023strategies, yoon2018strategies}. In reality, district heating systems  follow a structured topology structure in which supply and return lines exhibit inherent physical coupling, as illustrated in Fig.~\ref{fig:tree}. Graph-based learning provides a  natural representation of this structure by modeling sensors as nodes and  their physical or functional dependencies as edges \cite{niresi2024physics}. Spatial-temporal graph neural networks (STGNNs) \cite{jin2024survey} further extend this paradigm by capturing both the relational structures and the time-varying demand patterns. Recent deep learning studies in related spatial-temporal and energy-system domains further motivate the need for architectures that can explicitly handle multi-scale dependencies, heterogeneous variables, and incomplete measurements. For example, MGSFformer introduces a multi-granularity spatial-temporal fusion transformer for air-quality prediction, showing that fusing dependencies at different spatial and temporal resolutions can improve prediction performance in complex sensor networks \cite{yu2025mgsfformer}. GinAR+ addresses multivariate time-series forecasting under missing variables by jointly modeling interpolation and spatial-temporal dependencies, which is particularly relevant for infrastructure monitoring systems affected by sensor faults or communication gaps \cite{yu2025ginar+}. In energy applications, heterogeneous ensemble surrogate models have been used to support optimization of lithium-ion battery charging strategies, demonstrating the value of combining multiple data-driven models for nonlinear energy-system decision problems \cite{cheng2025charging}. Similarly, MetaGNSDformer combines meta-learning, gated non-stationary modeling, and frequency-aware attention for battery remaining useful life prediction, highlighting the importance of adaptive temporal representations under non-stationary operating conditions \cite{cheng2026metagnsdformer}. Collectively, these works highlight the growing importance of architectures that integrate heterogeneous information and complex temporal dynamics. However, they largely assume that all variables follow a common spatial-temporal dependency structure. This assumption is not valid for thermo-hydraulic networks, where flow, pressure, and temperature propagate according to fundamentally different physical mechanisms and timescales. Consequently, a key unresolved challenge is how to construct representations that explicitly account for variable-specific physical propagation while jointly exploiting their interactions for virtual sensing.  

Heterogeneous spatial–temporal graph neural networks have recently been explored in other cyber-physical systems where multiple sensing modalities interact. In domains such as intelligent transportation \cite{liang2022joint}, power systems \cite{li2023heterogeneous, theiler2025heterogeneous}, and air quality monitoring \cite{terroso2024nationwide, niresi2025efficient}, heterogeneous graph learning frameworks have been developed to jointly model heterogeneous measurements. These approaches typically exploit the complementary nature of different sensing sources to improve forecasting, anomaly detection, and state estimation. Despite these advances, the applicability of these methods to district heating networks remains relatively underexplored. Compared to transportation and electrical systems, district heating networks are governed by thermo-hydraulic processes that introduce additional temporal and spatial complexities. Measurements such as flow, temperature, and pressure are physically coupled through network constraints, which may affect how spatial and temporal dependencies should be modeled. At the same time, sensing infrastructures in operational district heating networks are often limited due to economic and practical considerations, leading to sparse, noisy, and unevenly distributed observations. These characteristics suggest that virtual sensing in this application requires models capable of capturing intra- and inter-variable interactions.  Another consideration is that many existing heterogeneous spatial–temporal GNN frameworks rely on application-specific graph construction strategies, often grounded in physical/spatial neighborhood  or domain knowledge, with graph structures that are manually specified or fixed a priori rather than learned from data \cite{zhao2025graph}. However, in district heating networks, such assumptions may not align well with the available sensing setup, since the relevant couplings between variables can depend on sensor placement \cite{niresi2025time}, operating conditions, and local thermo-hydraulic behavior. Consequently, the graph structure may need to be manually tailored to each deployment \cite{rodrigue2025validation, boghetti2023benchmark}, which limits the flexibility of these methods in data-sparse or structurally incomplete settings, particularly when the underlying network layout is unavailable, such as in legacy systems where documentation is missing.  Moreover, in district heating, different physical variables may exhibit distinct propagation mechanisms and temporal dynamics. For example, pressure variations can propagate rapidly through the network, at approximately the speed of sound in water (several km/s), whereas temperature dynamics evolve more gradually, with propagation delays that can exceed 60 minutes during summer, and depend on demand variability and thermal storage effects. This observation indicates that learning modality-specific spatial–temporal relationships could be beneficial for improving state estimation and virtual sensing performance in these systems.

To address these gaps, first, we propose an HSTGNN tailored to virtual sensing in district heating systems, where flow, temperature, and pressure exhibit distinct physical dynamics and coupling patterns. The model uses three branches to learn modality-specific graph structures and temporal dependencies, allowing each measurement type to be represented according to its own physical behavior. A dedicated fusion mechanism then combines these representations to infer unobserved sensor readings by exploiting both cross-variable dependencies and network-level spatial correlations. To support this study, we also introduce a controlled laboratory dataset collected at the Aalborg Smart Water Infrastructure Laboratory, providing synchronized high-resolution measurements of flow, temperature, and pressure under representative operating conditions. The dataset is designed to enable reproducible research and systematic benchmarking. Experiments demonstrate that the proposed method consistently outperforms baseline models in accuracy and robustness.

The main contributions of this work are:
\begin{itemize}
    \item A three-branch heterogeneous spatial-temporal graph neural network that learns dedicated graph structures and temporal dependencies for each sensor modality;
    \item A controlled laboratory dataset with synchronized flow, temperature, and pressure measurements, specifically designed for virtual sensing in district heating networks;
    \item A comprehensive evaluation demonstrating that the proposed framework consistently outperforms conventional machine learning methods, and homogeneous STGNN baselines.
\end{itemize}

Section \ref{sec:preliminaries} introduces the necessary preliminaries. Section \ref{sec:testbed} describes the testbed and dataset. Section \ref{sec:methodology} presents the proposed HSTGNN. Section \ref{sec:results} discusses the experimental results, and Section \ref{sec:conclusion} concludes by summarizing the main contributions and outlining directions for future work.

\section{Preliminaries}
\label{sec:preliminaries}
This section introduces the  concepts underlying  the proposed HSTGNN framework for virtual sensing in district heating networks. We first review Graph Neural Networks (GNNs) and their spatial-temporal extensions, which provide a flexible framework for modeling relational data and capturing both spatial and temporal dependencies. We then present domain-specific background, including typical sensor deployments and the physical relationships between flow, temperature, and pressure. These preliminaries establish the foundation  for designing data-driven models that infer unobserved variables and enable reliable monitoring in partially instrumented networks.

\subsection{Graph Deep Learning for Time Series}

In district heating networks, thermal-hydraulic interactions propagate through pipes and substations, creating coupling patterns that span multiple spatial scales. GNNs provide a powerful framework for learning over relational data, making them well suited for  physical systems in which sensor interactions are governed by  topology, spatial proximity, or domain-specific coupling. In a GNN, each node iteratively aggregates information from its neighbors through message-passing, allowing the model to capture localized spatial dependencies and multi-hop interactions. A general update for node $i$ at layer $l+1$ can be written as
\begin{equation}
\mathbf{h}_i^{(l+1)} = 
\sigma\left(
\mathbf{W}^{(l)} \mathbf{h}_i^{(l)} +
\sum_{j \in \mathcal{N}(i)}
f_\theta\!\left( \mathbf{h}_j^{(l)}, \mathbf{h}_i^{(l)}, \mathbf{e}_{ij} \right)
\right),
\end{equation}
where $\mathbf{h}_i^{(l)}$ is the node embedding at layer $l$, i.e., a vector of features describing node $i$, $\mathcal{N}(i)$ is the set of neighbors of node $i$, $\mathbf{e}_{ij}$ are edge attributes, and $f_\theta$ is a learnable message function \cite{kipf2017semisupervised, veličković2018graph}. This paradigm allows node representations to incorporate information from increasingly distant parts of the graph as more layers are stacked, yielding expressive models for tasks such as node regression, classification, and link prediction \cite{hamilton2020graph}.


STGNNs extend GNNs  to time-varying signals by jointly modeling spatial dependencies and  temporal dynamics. A widely used design is the time-then-graph architecture, in which  node-wise temporal sequences are first processed independently by a temporal encoder to extract latent representations, and spatial interactions are subsequently modeled through stacked GNN layers \cite{gao2022equivalence}. This separation allows the model to capture sensor-specific temporal patterns while leveraging the network structure to exchange information across physically connected components.

Formally, at each reference time step $t$, the model receives a temporal window of the past $T$ measurements for node $i$, namely $\mathbf{x}_i^{t-T+1:t} = \{x_{i,t-T+1}, \ldots, x_{i,t}\},$ where $t$ denotes the current time index and the notation $t-T+1:t$ indicates the inclusive interval of past observations. Unlike forecasting, where the temporal window includes only information available up to time $t$ and is used to predict future values, in virtual sensing the window is inclusive of time $t$, leveraging observations from available nodes at the current step to infer the values of unobserved ones. The temporal encoder maps this history to a latent representation,
\begin{equation}
\mathbf{h}_{i,t}^{(0)} = \mathrm{TemporalEncoder}\!\left(\mathbf{x}_i^{t-T+1:t}\right),
\end{equation}
which is then propagated through a sequence of graph neural network layers as
\begin{equation}
\mathbf{H}_{t}^{(l+1)} = \mathrm{GNN}^{(l)}\!\left(\mathbf{H}_{t}^{(l)}, \mathbf{A}\right), 
\quad l = 0, \ldots, L-1,
\end{equation}
where $\mathbf{A} \in \mathbb{R}^{N \times N}$ denotes the graph adjacency matrix and $\mathbf{H}_{t}^{(l)}$ collects the node embeddings at layer $l$. Common approaches include coupling GNN layers with recurrent networks such as Gated Recurrent Units (GRU) or Long Short-Term Memory (LSTM), or using temporal convolutional modules \cite{jin2024survey, wu2020comprehensive}.

\subsection{District Heating Networks Monitoring}

District heating networks employ three primary sensor types including flow, temperature, and pressure sensors to characterize the hydraulic and thermal state of the system. Flow and temperature measurements are fundamental for determining the thermal energy consumed at substations or delivered to specific network segments. Given the mass flow rate $\dot{m}$ and the supply and return temperatures $T_s$ and $T_r$, the instantaneous thermal power is computed as
\begin{equation}
\dot{q}= \dot{m} \, c_p \, (T_s - T_r),
\end{equation}
where $c_p$ denotes the specific heat capacity of water, typically treated as constant under standard district heating operating conditions. Smart heat meters implement  this relationship: they measure the mass flow rate $\dot{m}$ and the supply and return temperatures ($T_s$ and $T_r$) at regular intervals, compute instantaneous thermal power, and integrate it over time to obtain the accumulated heat energy up to time $t$:
\begin{equation}
Q(t) = \int_{\tau=0}^t \dot{q}(\tau)  d\tau,
\end{equation}
This forms the basis for consumer billing, performance monitoring, and energy accounting. Consequently, flow and temperature sensors directly measure key operational quantities such as heat demand, load profiles, and return temperature behavior.

In addition to these measurements,  pressure sensors play a crucial role in characterizing the  hydraulic state of the network. Their readings reflect not only local pressure drops or gradients but also encode latent information about network-wide flow dynamics, pump operations, and control strategies. As such, pressure measurements provide complementary signals that can improve the estimation of unobserved flow and temperature variables.

Despite these physical relationships, district heating networks are rarely fully instrumented  due to installation constraints, budget constraints, and the intrusive nature of pipe access. Consequently, sensing may be limited to a subset of nodes required for billing purposes. Virtual sensing therefore aims to infer missing sensor values by leveraging the spatial and temporal correlations among sensor types. Once reconstructed, these values can be used to estimate energy consumption at uninstrumented locations, supporting billing and operational decision-making.


\section{Dataset and Testbed}
\label{sec:testbed}

\subsection{Experimental Setups}

\subsubsection{Measurement Setup}
\begin{figure} \centering \includegraphics[width=0.7\linewidth]{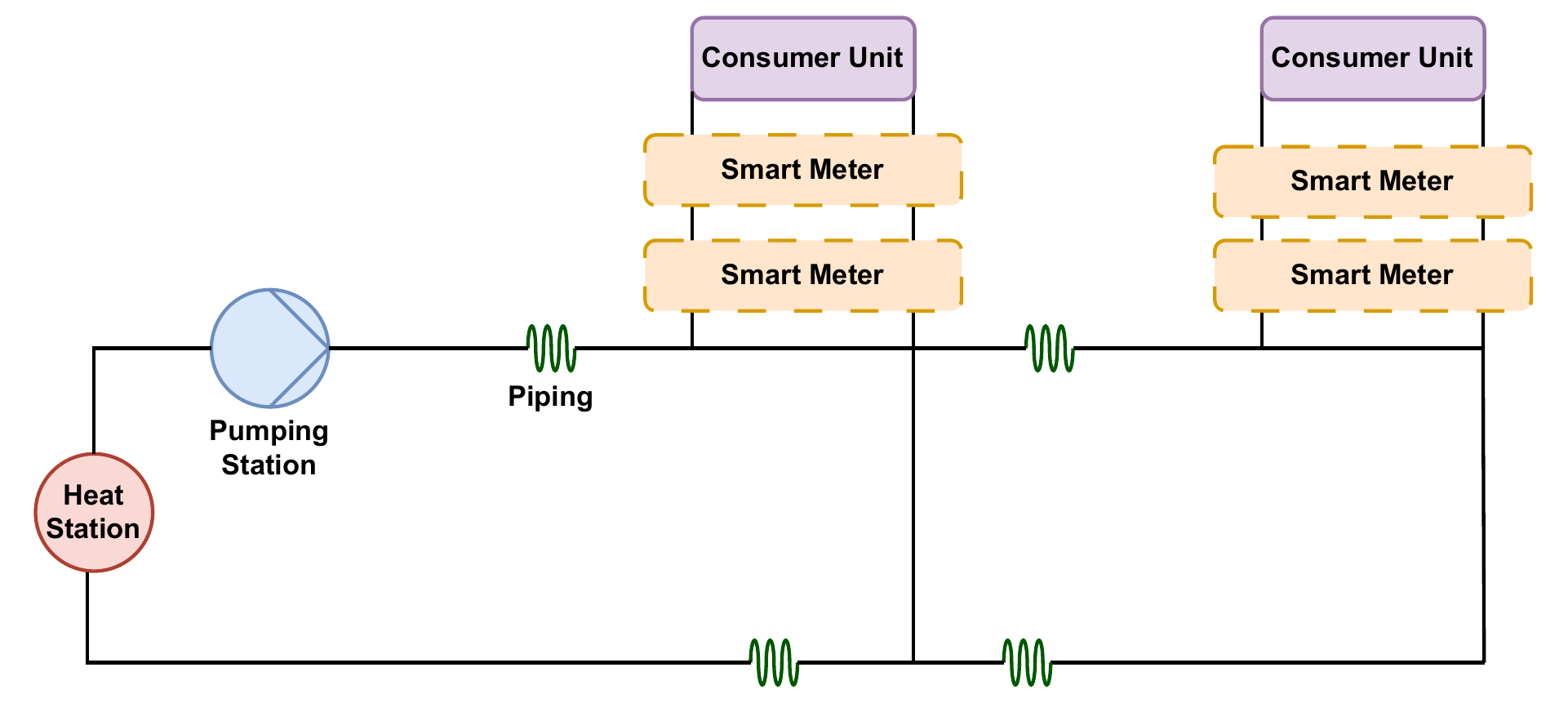} \caption{Schematic block diagram of the experimental topology. The system is connected in a tree structure, linking the heat station to consumers.} \label{diagram} 

\end{figure}

\begin{figure} \centering \includegraphics[width=0.5\linewidth]{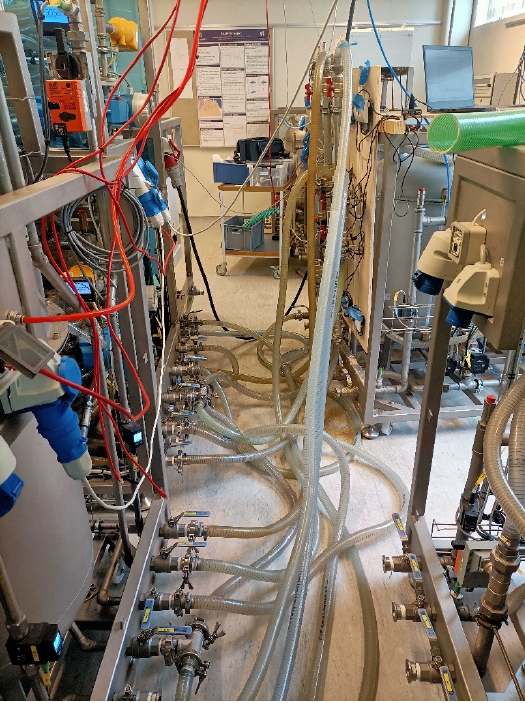} \caption{Hardware implementation of the experimental setup at Aalborg SWIL.} \label{setupfig} 

\end{figure}

The experimental data for this study were collected at the Smart Water Infrastructures Laboratory (SWIL) at Aalborg University, Denmark \cite{val2021smart}. SWIL is an experimental research facility designed for the simulation and analysis of water distribution and district heating systems under controlled condition. For this study, the modular infrastructure of the SWIL was utilized to emulate a representative district heating network. While the laboratory's flexible design allows for various structural and hydraulic configurations, this experiment specifically arranged the pumping units, heating modules, and interconnected piping into a tree-structured topology (Figure \ref{diagram}). This configuration  mirrors the supply and return pipeline layouts predominantly found in real-world municipal heating systems. The complete physical setup, which distributes thermal energy to two distinct consumer substations, is illustrated in Figure \ref{setupfig}.

To establish a highly reliable ground truth for model validation, rigorous instrumentation was applied at the network's endpoints. Each of the two consumer stations is equipped with two smart meters installed in series to measure the same physical quantities. The readings from these redundant sensors are averaged to minimize technical noise and ensure the accuracy of the reference measurements.

\begin{figure*} \centering \includegraphics[width=\linewidth]{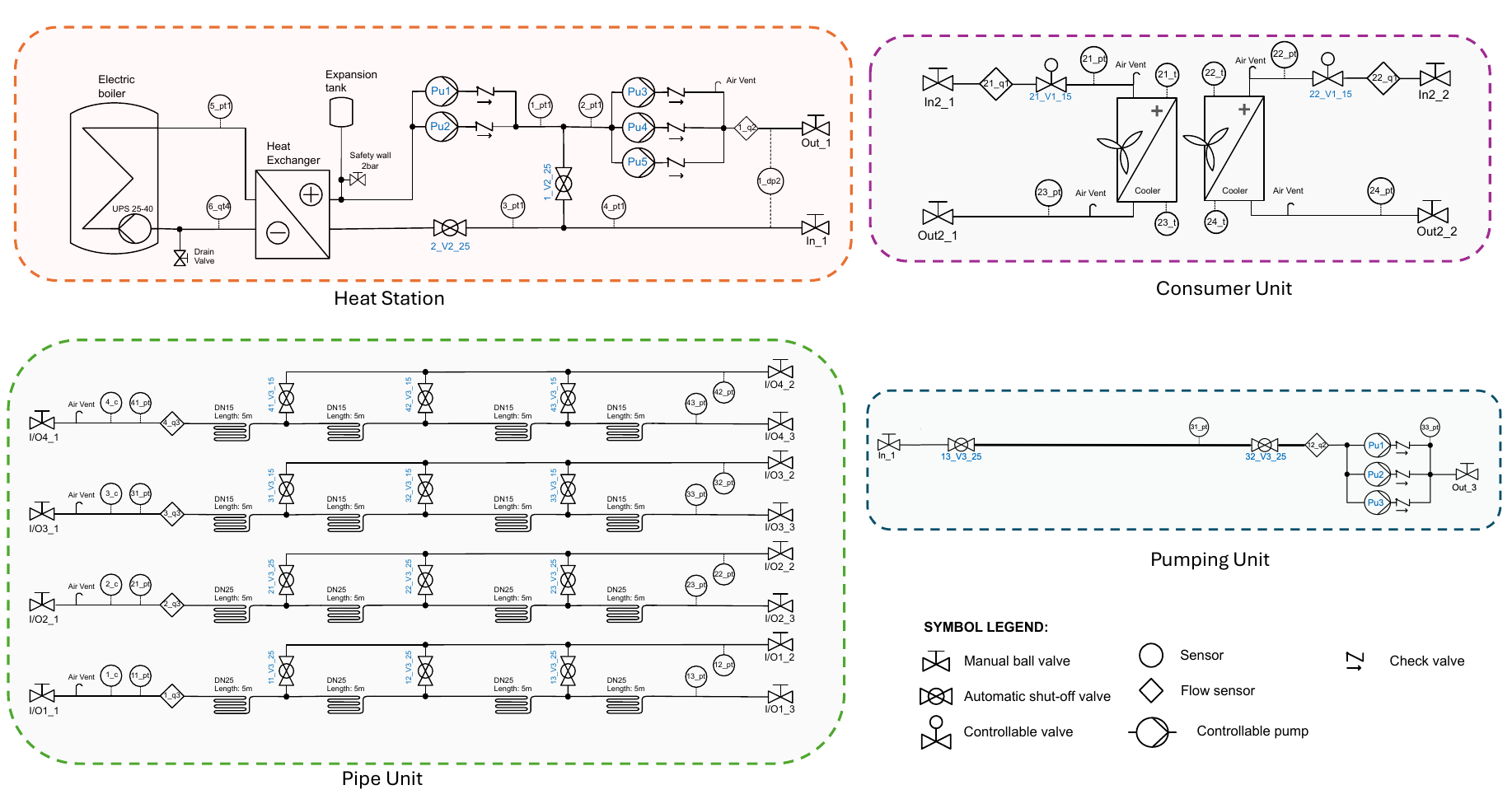} \caption{Modules at Aalborg Smart Water Infrastructures Laboratory. For the pipe units, the following input--output pairs are considered: 
(I/O1\_1, I/O1\_2), (I/O2\_1, I/O2\_3), (I/O3\_1, I/O3\_2), and (I/O4\_1, I/O4\_3).} \label{modules} 

\end{figure*}

This configuration was used to conduct controlled experiments under varying flow conditions, load patterns, and network states, providing realistic measurement data for evaluating the proposed monitoring approach. The internal configuration of the modules and the placement of sensors used in this study are illustrated in Fig.~\ref{modules}.

\subsubsection{Instrumentation and Data Acquisition}
To enable high-fidelity monitoring of the system states, the experimental setup was instrumented with industrial-grade sensors. Flow rates were measured using Endress \& Hauser Proline Promag 10 electromagnetic flow meters, while pressure dynamics were captured using Grundfos Direct Sensor RPI+T 0--1.6 units. Table \ref{tab:inputsensors} summarizes the input sensors used in the experiments. These measurements correspond to typical Supervisory Control and Data Acquisition (SCADA) signals and may be affected by measurement imperfections such as noise, miscalibration, or sensor drift. Furthermore, Kamstrup MULTICAL® 303 smart meters were deployed at consumer substations to emulate consumer-side heat metering points. The measurements obtained from these devices are considered high-accuracy reference signals and serve as ground truth for the study. These reference measurements inlet and outlet temperatures as well as flow rates at each consumer station.

\begin{table}[h]
\centering
\caption{Overview of the available system input sensors used as standard SCADA measurements}
\label{tab:inputsensors}
\resizebox{0.65\linewidth}{!}{%
\begin{tabular}{@{}llll@{}}
\toprule
\textbf{Component} & \textbf{Temperature} & \textbf{Pressure} & \textbf{Flow} \\
\midrule
\multirow{6}{*}{\textbf{Heat Station}}
& Boiler\_T1 & Boiler\_dP2 & Boiler\_Q2\_1 \\
& Boiler\_T2 & & Boiler\_Q4\_6 \\
& Boiler\_T3 & & \\
& Boiler\_T4 & & \\
& Boiler\_T5 & & \\
& Boiler\_T6 & & \\
\midrule
\multirow{8}{*}{\textbf{Pipe Unit}}
& Pipe\_T1\_1 & Pipe\_P1\_1 & Pipe\_Q3\_1 \\
& Pipe\_T1\_2 & Pipe\_P1\_2 & Pipe\_Q3\_2 \\
& Pipe\_T2\_1 & Pipe\_P2\_1 & Pipe\_Q3\_3 \\
& Pipe\_T2\_3 & Pipe\_P2\_3 & Pipe\_Q3\_4 \\
& Pipe\_T3\_1 & Pipe\_P3\_1 & \\
& Pipe\_T3\_2 & Pipe\_P3\_2 & \\
& Pipe\_T4\_1 & Pipe\_P4\_1 & \\
& Pipe\_T4\_3 & Pipe\_P4\_3 & \\
\midrule
\multirow{4}{*}{\textbf{Consumer Unit}}
& Consumer\_T\_21 & Consumer\_P\_21 & Consumer\_Q1\_21 \\
& Consumer\_T\_22 & Consumer\_P\_22 & Consumer\_Q1\_22 \\
& Consumer\_T\_23 & Consumer\_P\_23 & \\
& Consumer\_T\_24 & Consumer\_P\_24 & \\
\midrule
\multirow{2}{*}{\textbf{Pumping Station}}
& Pump\_T3\_1 & Pump\_P3\_1 & Pump\_Q2\_12 \\
& Pump\_T3\_3 & Pump\_P3\_3 & \\
\bottomrule
\end{tabular}%
}
\end{table}

The data acquisition and control architecture follows a hierarchical structure. Sensor measurements are collected through Beckhoff I/O modules, which interface with Raspberry Pi units running CodeSys soft Programmable Logic Controllers (PLCs) for local module control. The data are transmitted to the Central Control Unit (CCU) via the Modbus Transmission Control Protocol (TCP) protocol. High-level orchestration and data logging are performed using the MATLAB Industrial Communications Toolbox, running in the MATLAB Simulink environment. To ensure strict temporal synchronization across all distributed sensors, the system employs the Simulink Real-Time Pacer. All sensor signals are resampled at a uniform rate of one measurement every two seconds (0.5~Hz), ensuring consistent temporal resolution across all sensor types.

\subsubsection{Smart Meter Operation and Data Acquisition Constraints}
A total of approximately 30 hours of synchronized operational data (approximately 54000 datapoints) were collected from the experimental setup. During data acquisition, an important practical constraint was observed in the operation of commercial smart meters.  To conserve battery life, these devices typically enter a sleep mode after approximately five minutes of continuous operation. To enable extended measurements during the experiments, the meters were configured in a dedicated ``test loop'' mode, which allows continuous operation for up to nine hours before requiring a reset.


\subsection{Datasets}
The datasets used in this study were collected during four distinct operating periods of the district heating network at the SWIL. To emulate changes in consumer demand, the fan speed at each consumer station was adjusted between periods. Specifically, the four operating conditions were defined by different combinations of fan speeds: (i) both consumers operating at maximum speed, (ii) both consumers operating one step below maximum speed, (iii) one consumer operating at maximum speed while the other operated one step below maximum speed, and (iv) the reverse of the previous case. In addition, a control strategy was applied to introduce further variability in the thermal conditions. When the boiler temperature decreased, leading to a drop in the supply water temperature, the boiler was activated to restore the supply temperature to approximately 50 °C. Once this target temperature was reached, the boiler was switched off again. This on–off control mechanism introduced additional temporal variations in the temperature profiles, thereby enriching the datasets with more diverse operating conditions. These operating conditions induced different hydraulic and thermal responses in the network, resulting in four datasets that capture a range of realistic operating regimes. Table \ref{tab:system_ranges} summarizes the operating ranges for temperature, flow, and pressure, defined by the 5th and 95th quantiles across all relevant sensors.
Prior to each data collection period, the system was operated for approximately 30 minutes to allow  thermal and hydraulic conditions to stabilize. This warm-up phase minimizes the influence of transient effects and ensures that the recorded measurements reflect steady operating conditions.

\begin{table}[h]
\centering
\caption{System Operating Ranges (5th – 95th Quantile)}
\label{tab:system_ranges}
\begin{tabular}{@{}lll@{}}
\toprule
\textbf{Variable} & \textbf{Range (5th – 95th)} & \textbf{Unit} \\ \midrule
Temperature   & 16.89 – 49.93                       & °C            \\
Flow    & 0.09 – 12.59                        & lit/min         \\
Pressure       & 3.27 – 10.71                        & bar           \\ \bottomrule
\end{tabular}
\end{table}

Publicly available district heating datasets remain scarce, particularly for virtual sensing tasks that require synchronized multi-physics measurements. Existing public resources, such as \cite{schaffer2022three}, provide valuable long-term hourly measurements from thousands of residential heat meters and are well suited for load analysis, demand characterization, and building-level heat consumption studies. However, such datasets typically focus on consumer smart meter records and do not provide the synchronized internal SCADA measurements of flow, pressure, and temperature needed to evaluate network-level virtual sensing models. In contrast, the dataset introduced in this study provides high-resolution synchronized measurements from multiple sensor types, including flow, pressure, and temperature, together with consumer-side smart meter reference signals. Although the laboratory network is smaller than an operational municipal system, the dataset fills a complementary benchmarking gap by enabling controlled and reproducible evaluation of multi-physics virtual sensing methods under known operating conditions. The dataset is publicly available on Zenodo at
\href{https://doi.org/10.5281/zenodo.23127439}
{https://doi.org/10.5281/zenodo.23127439}.

\subsection{Data Preprocessing}

Each dataset is evaluated using a leave-one-dataset-out strategy to assess generalization across different operating conditions. In each experiment, three datasets are used for training, while the remaining dataset is reserved for inference. For example, Datasets~1, 2, and~3 are used for training and Dataset~4 is used for inference. This procedure is repeated by rotating the inference dataset so that each operating condition is evaluated independently. This evaluation protocol ensures that the model is tested on operating regimes that are not observed during training.

Feature normalization is performed using statistics computed exclusively from the training datasets in order to prevent information leakage. Each input signal is standardized according to
\begin{equation}
\mathbf{X}_{\mathrm{norm}} = \frac{\mathbf{X} - \boldsymbol{\mu}_{\mathrm{train}}}{\boldsymbol{\sigma}_{\mathrm{train}}},
\end{equation}
where $\mathbf{X}$ denotes the original signal, $\boldsymbol{\mu}_{\mathrm{train}}$ and $\boldsymbol{\sigma}_{\mathrm{train}}$ represent the mean and standard deviation of the training data, respectively, and $\mathbf{X}_{\mathrm{norm}}$ is the normalized signal.

To prepare the data for temporal modeling, a sliding-window approach is applied to the normalized signals.

\section{Methodology}
\label{sec:methodology}

In this work, we propose a model architecture tailored for virtual sensing in district heating networks. The objective is to infer unobserved sensor measurements by leveraging the  spatial and temporal dependencies inherent in the system. Unlike conventional approaches that treat all sensor signals uniformly, the proposed method explicitly accounts for the heterogeneous nature of physical sensors, namely temperature, pressure, and flow measurements, through type-specific processing branches. Each branch captures the  temporal dynamics and intra-type spatial relationships associated with  its corresponding sensor type. Interactions across different sensor types are modeled using an attention mechanism, enabling the model to learn how different physical quantities influence one another.

\subsection{Problem Formulation}

Let $\mathbf{X} \in \mathbb{R}^{N \times T}$ denote a multivariate time-series input with window length $T$ and $N$ sensor nodes. Each node $v_i \in \mathcal{V}$, for $i = 1, \ldots, N$, is associated with a univariate time-series corresponding to a single physical measurement. The objective is to predict a target vector $\hat{\mathbf{y}} \in \mathbb{R}^{D}$, where $D$ denotes the number of virtual sensors to be estimated.

The node set $\mathcal{V}$ is partitioned into three disjoint subsets according to sensor type. Temperature sensors are denoted by $\mathcal{V}_{\text{temp}}$, pressure sensors by $\mathcal{V}_{\text{press}}$, and flow sensors by $\mathcal{V}_{\text{flow}}$, with respective cardinalities $N_{\text{temp}}$, $N_{\text{press}}$, and $N_{\text{flow}}$. The total number of sensor nodes satisfies
\begin{equation}
N = N_{\text{temp}} + N_{\text{press}} + N_{\text{flow}}.
\end{equation}

At time step $t$, the measurements from each sensor type are represented as
\begin{equation}
\mathbf{x}_{\text{temp}}^{t} \in \mathbb{R}^{N_{\text{temp}}}, \quad
\mathbf{x}_{\text{press}}^{t} \in \mathbb{R}^{N_{\text{press}}}, \quad
\mathbf{x}_{\text{flow}}^{t} \in \mathbb{R}^{N_{\text{flow}}},
\end{equation}
where each vector collects the measurements of all sensors belonging to the corresponding sensor type. To capture temporal dependencies, sliding windows of length $T$ are constructed for each sensor type:
\begin{align}
\mathbf{X}_{\text{temp}}^{t-T+1:t} &= 
\big[
\mathbf{x}_{\text{temp}}^{t-T+1}, \ldots, \mathbf{x}_{\text{temp}}^{t-1}, \mathbf{x}_{\text{temp}}^{t}
\big]
\in \mathbb{R}^{N_{\text{temp}} \times T}, \\
\mathbf{X}_{\text{press}}^{t-T+1:t} &= 
\big[
\mathbf{x}_{\text{press}}^{t-T+1}, \ldots, \mathbf{x}_{\text{press}}^{t-1}, \mathbf{x}_{\text{press}}^{t}
\big]
\in \mathbb{R}^{N_{\text{press}} \times T}, \\
\mathbf{X}_{\text{flow}}^{t-T+1:t} &= 
\big[
\mathbf{x}_{\text{flow}}^{t-T+1}, \ldots, \mathbf{x}_{\text{flow}}^{t-1}, \mathbf{x}_{\text{flow}}^{t}
\big]
\in \mathbb{R}^{N_{\text{flow}} \times T}.
\end{align}

The virtual sensing task is to learn a mapping
\begin{equation}
f\!\left(
\mathbf{X}_{\text{temp}}^{t-T+1:t},
\mathbf{X}_{\text{press}}^{t-T+1:t},
\mathbf{X}_{\text{flow}}^{t-T+1:t}
\right)
= \hat{\mathbf{y}}_t,
\end{equation}
that estimates the target variables $\hat{\mathbf{y}}_t \in \mathbb{R}^{D}$ at time $t$, corresponding to unobserved or unavailable sensor measurements, such as consumer-side smart meter readings. This task is challenging due to the heterogeneous temporal dynamics of the sensor types and their coupled thermal--hydraulic relationships under varying operating conditions.

\subsection{Graph Representation of Multi-Sensor Networks}

In the proposed formulation, each node in the graph corresponds to a single physical sensor. The sensor network is represented as a graph:
\begin{equation}
\mathcal{G} = (\mathcal{V}, \mathcal{E}, \mathbf{A}),
\end{equation}
where $\mathcal{V} = \{v_1, v_2, \dots, v_{|\mathcal{V}|}\}$ denotes the set of sensors, $\mathcal{E}$ denotes sensor-to-sensor relationships, and $\mathbf{A} \in \mathbb{R}^{|\mathcal{V}| \times |\mathcal{V}|}$ is the adjacency matrix encoding edge weights. Each node $v_i$ measures a single physical quantity, such as flow rate, temperature, or pressure. To make this explicit, we define a sensor type mapping:
\begin{equation}
\tau: \mathcal{V} \to \{\mathrm{F}, \mathrm{T}, \mathrm{P}\},
\end{equation}
where $\tau(v_i)$ indicates whether sensor $v_i$ is a flow (F), temperature (T), or pressure (P) sensor.

Since each sensor measures a single quantity, the instantaneous measurement at node $v_i$ is a scalar denoted $x_i \in \mathbb{R}$. The instantaneous network state is therefore the vector:
\begin{equation}
\mathbf{x} =
\begin{bmatrix}
x_1 \\
\vdots \\
x_{|\mathcal{V}|}
\end{bmatrix}
\in \mathbb{R}^{|\mathcal{V}|},
\end{equation}
together with the sensor type labels $\{\tau(v_i)\}_{i=1}^{|\mathcal{V}|}$. For modeling convenience, sensor type information can be encoded as auxiliary categorical or one-hot vectors associated with each node, or as separate input branches that process only nodes of a given sensor type.

Edges in $\mathcal{E}$ capture sensor-to-sensor correlations. Edge weights can be constructed from physical metrics, such as pipe distance or hydraulic connectivity, or learned from data to represent latent influence between sensors. This sensor-as-node representation enables the model to learn sensor-to-sensor relationships, including cross-type interactions.

\begin{figure*}
    \centering
    \includegraphics[width=\linewidth]{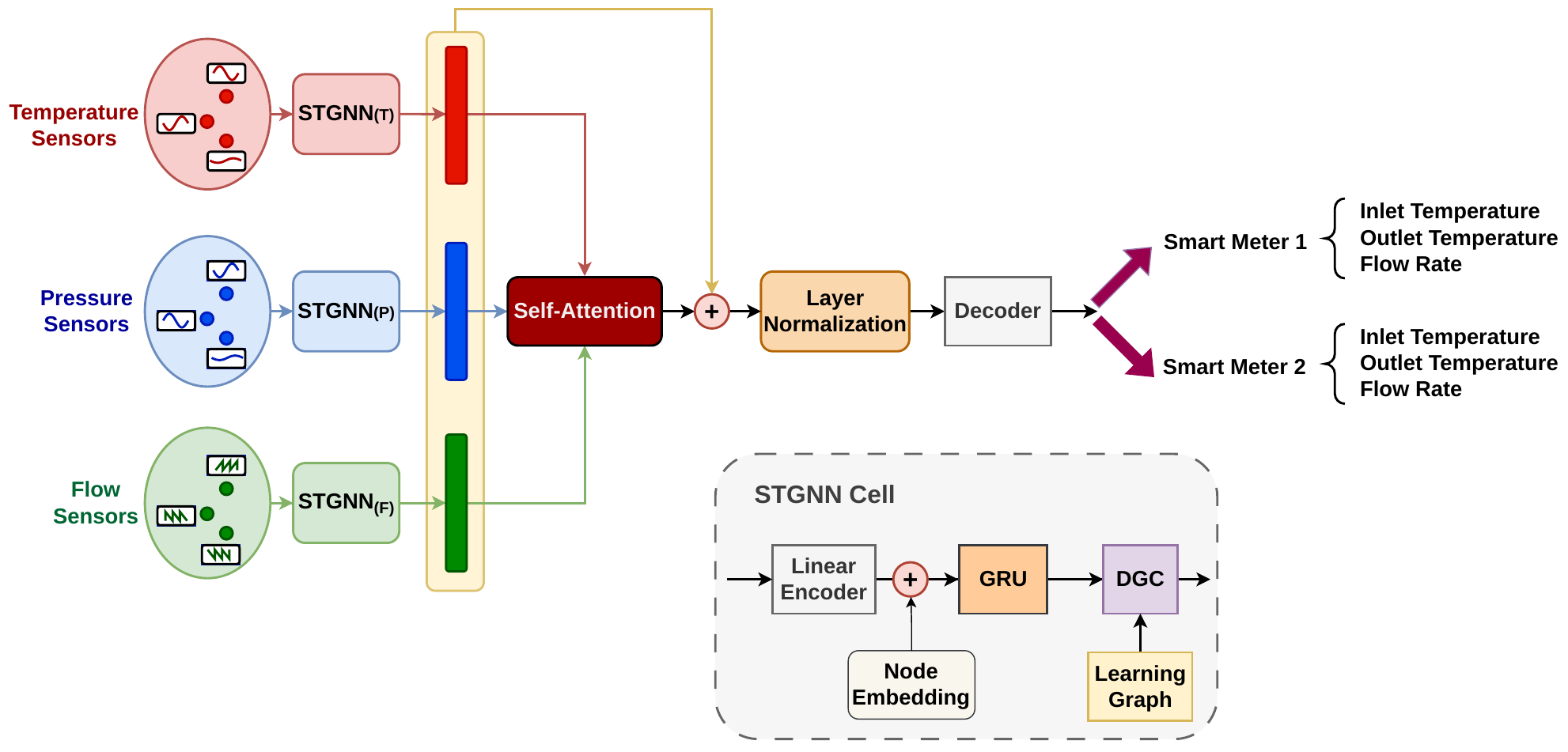}
    \caption{The Proposed HSTGNN Architecture for Virtual Sensing}
    \label{fig:HSTGNN}
\end{figure*}

\subsection{Model Architecture}

The proposed architecture is a multi-branch STGNN designed to explicitly account for heterogeneous \emph{sensor types}. Each type of physical sensor is processed by a dedicated branch with a shared architectural structure but independent parameters. This design is motivated by the distinct physical dynamics and measurement characteristics of different sensor types, such as flow, temperature, and pressure, which exhibit unique temporal behaviors and spatial dependencies across the network. The model can capture these type-specific patterns before integrating information across sensor types by assigning each sensor type to its own processing branch.

Each branch follows the same processing pipeline: (i) input encoding with sensor-specific information, (ii) temporal modeling at the sensor level, and (iii) spatial modeling using a learned graph. The outputs of all branches are then integrated through a cross-type attention mechanism, which allows the model to selectively fuse information from different sensor modalities, and decoded for final prediction, as shown in Fig. \ref{fig:HSTGNN}. This multi-branch design ensures that the heterogeneous nature of the sensors is explicitly respected while still enabling a unified global representation for virtual sensing.

\subsubsection{Input Encoding and Sensor-Specific Embedding}

Each sensor provides a univariate time series. At each time step, measurements are first projected into a higher-dimensional hidden space using a linear transformation. This step will increase the representational capacity of the model. The core intuition behind this stage is twofold: first, to capture the distinct characteristics associated with different physical variables, and second, to preserve the identity of individual sensors. Although sensors may share the same physical type, their behavior can differ depending on their specific location within the district heating network. To explicitly encode sensor identity, each branch includes a learnable node embedding associated with every sensor of that type. These embeddings are added to the projected features and allow the model to distinguish between sensors even when their measurement patterns are similar and help contextualize each node within the broader network.

Let the input of sensor $v_i$ be denoted as $ \mathbf{x}_i \in \mathbb{R}^{T \times 1}.$
After linear projection and addition of the node embedding, the encoded sequence is
\begin{equation}
\mathbf{Z}_i = \mathrm{Linear}_{\tau(v_i)}(\mathbf{x}_i) + \mathbf{1}_T \mathbf{e}_i,
\end{equation}
where $\mathbf{Z}_i \in \mathbb{R}^{T \times d}$, $\mathbf{e}_i \in \mathbb{R}^{d}$ is the learnable embedding associated with sensor $i$, and $\mathbf{1}_T$ replicates the embedding across all time steps.

\subsubsection{Temporal Modeling}

Temporal dependencies are modeled for each sensor using Gated Recurrent Units (GRUs). Sensors that measure the same physical quantity share the same GRU parameters, resulting in one GRU per sensor type (e.g., temperature, pressure, and flow). Each GRU processes the full temporal window of an individual sensor and produces a fixed-length representation summarizing its recent dynamics:

\begin{equation}
\mathbf{h}_i = \mathrm{GRU}(\mathbf{Z}_i; \Theta_{\tau(v_i)}),
\end{equation}
where $\Theta_{\tau(v_i)}$ denotes the GRU parameters associated with the physical type $\tau(v_i)$ of sensor $v_i$. Consequently, sensors of the same type share temporal modeling parameters while still producing independent representations based on their individual measurement histories. Only the final hidden state of each GRU is retained. This design choice reduces computational complexity and allows the spatial component of the model to operate on temporally aggregated representations rather than raw sequences. Stacking all node representations of type $m$ yields $ \mathbf{H}^{(m)} \in \mathbb{R}^{N_m \times d_h},$
where each row corresponds to $\mathbf{h}_i$. Each node representation already captures its local temporal behavior prior to interacting with neighboring sensors, by applying temporal modeling before spatial aggregation.

\subsubsection{Learning the Graph Structure}

Rather than relying on a predefined physical topology, the spatial structure within each sensor type is learned directly from data. Assuming the dependency structure among the time series is unknown, we associate each possible edge within a sensor type with a learnable score $\phi_{ij}$, where $i,j \in \{1, \dots, N_m\}$ and $N_m$ denotes the number of nodes in sensor type $m$ (e.g., temperature, pressure, or flow).  
For each node $i$, a subset of $K$ neighbors is sampled without replacement according to the categorical distribution $\boldsymbol{\pi}_i^{(m)}$, whose entries are defined as \cite{niresi2026relational, cini2023sparse, cini2025relational}
\begin{equation}
\pi_{ik}^{(m)} =
\frac{\exp(\phi_{ik}^{(m)})}{\sum_{j=1}^{N_m} \exp(\phi_{ij}^{(m)})},
\quad k = 1, \dots, N_m.
\end{equation}

The neighbor set $M_i$ is then obtained by sampling $K$ distinct nodes from this distribution. Here, $\Phi^{(m)}$ contains directly learnable parameters for branch $m$, and $M_i$ denotes the set of $K$ sampled neighbors of node $i$, defining a sparse and learned graph structure that is used in the subsequent GNN message-passing layers. Sampling is implemented efficiently using the GumbelTop$K$ trick \cite{kool2020ancestral} with a straight-through gradient estimator \cite{bengio2013estimating}. In the forward pass, discrete neighbor selections are used, while in the backward pass a surrogate gradient is used to update the score parameters.

In practical district heating deployments, the graph structure can be constructed in different ways depending on the availability and reliability of physical network information. When detailed topology, pipe lengths, valve states, or flow directions are available, these data can be used to define an initial adjacency matrix or to constrain the candidate neighbor set. However, such information is often incomplete in legacy networks, and the hydraulic connectivity observed by sensors may deviate from the nominal pipe layout because of control actions, shared pumps, valve settings, and time-varying operating conditions. For this reason, the proposed HSTGNN learns a separate score matrix $\Phi^{(m)}$ for each sensor type directly from historical operational data. Each entry $\phi_{ij}^{(m)}$ represents the learnable strength of a functional dependency between two sensors of the same type, and the GumbelTopK sampling mechanism selects the most informative neighbors during training. Thus, the learned graph should be interpreted as a functional dependency graph rather than only as a pipe-connectivity graph. If physical topology is available, it can be incorporated as an initialization, mask, or regularization term; if it is not available, the model can still infer data-driven dependencies from synchronized sensor measurements. This makes the approach suitable for real-world district heating systems where documentation is incomplete or where relevant dependencies are induced by both physical layout and operational control.

\subsubsection{Spatial Modeling}

Spatial dependencies are modeled using a diffusion-based graph convolution (DGC), which propagates information across the learned graph and allows each sensor to incorporate information from its neighbors. Limiting the graph convolution to within-type nodes helps reduce noise and promotes physically meaningful interactions. Formally, the diffusion convolution is defined as \cite{atwood2016diffusion, li2018diffusion}:
\begin{equation}
\mathbf{H}_{\text{sp}}^{(m)} =
\psi\!\left(
\sum_{s=0}^{S} \mathbf{P}^{(m)s} \mathbf{H}^{(m)} \mathbf{W}^{(s)}
\right),
\end{equation}
where $\mathbf{P}^{(m)} = \mathbf{D}^{(m)-1}\mathbf{A}^{(m)}$ is the normalized transition matrix, $\mathbf{H}^{(m)}$ contains the node features (output of the temporal module), $\mathbf{W}^{(s)}$ are learnable weights, $\psi$ is a nonlinear activation, and $S$ is the number of diffusion steps. This operation captures information from multi-hop neighborhoods, enriching each node’s spatial representation. Through successive powers of the transition matrix, this operation enables information to propagate over multi-hop neighborhoods, thereby enriching each node’s spatial representation with both local and higher-order structural information. For directed graphs, the transition matrix diffusion along the direction of edges, allowing each node to aggregate information from its outgoing neighbors across multiple steps. However, spatial dependencies in real-world systems are often not strictly unidirectional. To capture bidirectional interactions, diffusion can also be performed over the transposed connectivity, which propagates information along incoming edges. Incorporating both forward and reverse diffusion allows each node to account for influences from both downstream and upstream neighbors, leading to more comprehensive and expressive spatial representations in directed settings.

\subsubsection{Cross-Type Interaction via Attention}

After intra-type processing, all sensor nodes are combined to form a single, unified graph in which every node can interact with nodes of other sensor types. To capture cross-type dependencies, the model applies a self-attention mechanism across this unified graph \cite{vaswani2017attention}. This attention layer allows each sensor to selectively attend to other sensors, regardless of type, learning interactions such as how pressure and flow sensors influence one another. A residual connection and layer normalization are applied to stabilize training and preserve the original node features.

\subsubsection{Decoding}

After the attention mechanism, all sensor representations are aggregated into a single feature vector and passed through a linear decoder to produce the predictions. This approach ensures that each output depends on the combined information from all sensors, allowing the model to capture inter-sensor dependencies and deliver more virtual sensing for heterogeneous sensor types.

\subsection{Training and Inference}

The HSTGNN model is trained to estimate consumer-side smart meter readings (three sensors for each smart meter) simultaneously using data from the set of monitored sensors, which may be incomplete or subject to noise. During training, the network receives input time series from the available sensors and learns to predict the target measurements at smart meters installed at consumer endpoints. The training objective is the mean absolute error, chosen for its robustness to outliers and measurement noise common in real-world district heating networks. During inference, only the monitored sensor readings are available. The trained network propagates information through the type-specific branches, models latent spatial relationships, and applies cross-type attention to generate predictions for all smart meters.

\section{Experimental Results}
\label{sec:results}

\subsection{Performance Analysis}

We evaluate the performance of the proposed HSTGNN model against several baseline methods, including GRU-GCN, LSTM, 1D-CNN, DGC, and GCN, selected to represent a range of modeling capabilities. In particular, LSTM capture temporal dependencies, 1D-CNN provides a convolutional temporal baseline, GCN and DGC model spatial relationships through graph-based operations, and GRU-GCN combines both spatial and temporal components. This selection enables a comprehensive comparison between purely temporal, purely spatial, and spatial-temporal approaches. The evaluation is conducted across multiple datasets and smart meters, considering both flow rate and temperature variables, which exhibit different temporal characteristics and levels of spatial dependency. Quantitative results are reported using standard regression error metrics, while qualitative visualizations illustrate the ability of each model to track sensor dynamics over time.

To measure prediction accuracy, we employ the Root Mean Squared Error (RMSE) and the Mean Absolute Error (MAE), defined as

\begin{equation}
\text{RMSE} = 
\sqrt{
\frac{1}{M}
\sum_{i=1}^{M}
\left( \hat{y}_i - y_i \right)^2
},
\end{equation}

\begin{equation}
\text{MAE} =
\frac{1}{M}
\sum_{i=1}^{M}
\left| \hat{y}_i - y_i \right|,
\end{equation}
where \(M\) denotes the total number of evaluation samples, \(y_i\) is the ground-truth sensor measurement, and \(\hat{y}_i\) is the corresponding model prediction. RMSE places greater emphasis on large deviations, whereas MAE provides a more interpretable measure of average absolute error.

\subsection{Hyperparameters}
Model hyperparameters and data preprocessing configurations were determined via a systematic grid search over a predefined parameter space. To maintain experimental integrity and prevent information leakage, all parameters were optimized using the first dataset; these configurations were subsequently held constant across all remaining datasets. This approach ensures that performance variations are attributable to the models' generalization capabilities rather than dataset-specific fine-tuning. For temporal preprocessing, window sizes of $\{4, 8, 16, 32, 64\}$ were evaluated, with a window size of $16$ selected for providing the optimal balance between predictive accuracy and computational overhead. A uniform stride of $1$ was applied across all experiments. The investigated search space and the resulting optimal configurations—highlighted in bold—are summarized in Table \ref{tab:hyperparameters}.

For the baseline models, we adopt the same procedure to ensure a fair comparison. The LSTM model employs a hidden dimension of $256$ with $2$ stacked recurrent layers, followed by a fully connected projection to the output space. The 1D-CNN baseline uses a single convolutional layer with $64$ filters, followed by adaptive pooling and a linear layer mapping from the $64$-dimensional feature space to the output. For graph-based baselines, all models share a latent node representation size of $16$. Node features are first projected into this hidden space and combined with learnable node embeddings of the same dimensionality. The DGC model applies a diffusion convolution with two propagation step over the learned graph, while the GCN baseline employs a standard graph convolution operating on the same $16$-dimensional node features. In both cases, the resulting node representations are passed through a linear decoder. To incorporate temporal dynamics, the GRU-GCN model introduces a temporal encoder consisting of a $2$-layer GRU with hidden size $16$, producing node-level embeddings that are subsequently processed by the graph convolution layer.

\begin{table}
\centering
\caption{Summary of the hyperparameter search space and optimal configurations. Values selected via grid search on the primary dataset and utilized across all experiments are indicated in bold.}
\resizebox{0.45\linewidth}{!}{%
\begin{tabular}{ll}
\toprule
\textbf{Hyperparameter} & \textbf{Value} \\ \midrule\midrule

Window Size & $\{4, 8, \textbf{16}, 32, 64\}$ \\ \midrule
Stride & $1$ \\ \midrule
Number of GNN Layers &  $\{\textbf{1}, 2, 3, 4\}$ \\ \midrule
Output Channels for GNN Layers & $\{4, 8, \textbf{16}, 32, 64\}$ \\ \midrule
GNN Kernel Size & $\{1, \textbf{2}, 3, 4\}$  \\ \midrule
Node Embedding Size & $\{4, 8, \textbf{16}, 32, 64\}$ \\ \midrule
$K$ (for Learning Graph) & $\{2, 4, 6, \textbf{8}, 10, 12\}$ \\ \midrule
Number of GRU Layers & $\{1, \textbf{2}, 3, 4\}$  \\ \midrule
Hidden Size for GRU Layers & $\{4, 8, \textbf{16}, 32, 64\}$ \\ \midrule
Batch Size & 512 \\ \midrule
Loss & Mean Absolute Error \\ \midrule
Optimizer & Adam \\ \midrule
Learning Rate & 0.01 \\ \bottomrule
\end{tabular}
}
\label{tab:hyperparameters}
\end{table}

\subsection{Quantitative Results}

Table~\ref{tab:dataset1_opt} summarizes the performance on Dataset~1 for flow rate, inlet temperature, and outlet temperature across two smart meters (SMs). The results demonstrate that methodologies combining temporal modeling with graph-based spatial modeling (GRU-GCN and HSTGNN) outperform other baselines. The proposed HSTGNN exhibits superior performance for thermodynamic variables, achieving the lowest errors for both inlet and outlet temperatures. While LSTM achieves a marginally lower RMSE for flow rate on SM~1, HSTGNN remains highly competitive and yields the optimal MAE. In contrast, purely spatial baselines (GCN and DGC) show severe performance degradation on temperature sensors, indicating that spatial topology alone is insufficient to capture highly dynamic transient behaviors.

These observations are supported by the results from Dataset~2 (Table~\ref{tab:dataset2_opt}), where the advantages of the proposed spatial-temporal architecture are most pronounced for temperature prediction. HSTGNN achieves the highest accuracy and stability for both inlet and outlet temperatures across all smart meters. For flow rate prediction, GRU-GCN and GCN demonstrate marginal advantages for SM~1 and SM~2, respectively. However, the performance gap between the leading methods on SM~2 is negligible. This suggests that while spatial dependencies may dominate specific sensor modalities under certain conditions, HSTGNN effectively balances these heterogeneous requirements without sacrificing overall generalizability.

Table~\ref{tab:dataset3_opt} presents the evaluation on Dataset~3, a regime where performance disparities between the evaluated methods become significantly more evident. Under this condition, HSTGNN demonstrates exceptional robustness, consistently securing the lowest RMSE across nearly all sensors and smart meters. While GRU-GCN remains competitive in minimizing absolute errors (MAE) for flow rate estimation, sequence-only baseline models (LSTM, 1D-CNN) exhibit a marked decrease in accuracy. This degradation highlights their limited capacity to leverage the crucial inter-sensor spatial correlations required to resolve complex and fluctuating operational dynamics.

The empirical evaluation on Dataset~4 (Table~\ref{tab:dataset4_opt}) definitively establishes the superiority of the proposed architecture. HSTGNN achieves a comprehensive sweep, leading to the lowest RMSE and MAE across all variables and smart meters.

Synthesizing the empirical findings across all four datasets (Tables~\ref{tab:dataset1_opt}–\ref{tab:dataset4_opt}), it is evident that frameworks jointly modeling temporal dynamics and latent spatial interactions yield superior performance. HSTGNN’s consistently dominant results, particularly on challenging temperature sequences, strongly validate our architectural design choices. Type-specific processing allows the network to isolate sensor-dependent dynamics, while the learned graph structures flexibly capture underlying spatial dependencies. Ultimately, the cross-type attention mechanism facilitates critical information routing between heterogeneous physical quantities, leading to highly accurate predictive modeling.

\begin{table*}[ht]
\centering
\caption{Dataset 1: RMSE and MAE comparison across methods (mean $\pm$ std over 3 runs)}
\label{tab:dataset1_opt}
\resizebox{\linewidth}{!}{%
\begin{tabular}{lllcccccc}
\toprule
\multirow{2}{*}{} & \multirow{2}{*}{Sensor} & \multirow{2}{*}{Metric} 
& \multicolumn{6}{c}{Methods} \\
\cmidrule(lr){4-9}
& & & GRU-GCN & LSTM & 1D-CNN & DGC & GCN & HSTGNN \\
\midrule
\multirow{6}{*}{SM1} 
& \multirow{2}{*}{Flow Rate ($\mathrm{l/min}$)} 
  & RMSE & 0.1969 $\pm$ 0.0017 & \textbf{0.1749 $\pm$ 0.0011} & 0.2416 $\pm$ 0.0022 & 0.1934 $\pm$ 0.0007 & 0.1940 $\pm$ 0.0013 & 0.1856 $\pm$ 0.0009 \\
& & MAE  & 0.1157 $\pm$ 0.0035 & 0.1062 $\pm$ 0.0017 & 0.1561 $\pm$ 0.0006 & 0.1170 $\pm$ 0.0009 & 0.1180 $\pm$ 0.0026 & \textbf{0.1017 $\pm$ 0.0015} \\
\cmidrule(lr){2-9}
& \multirow{2}{*}{Inlet Temperature ($^\circ\mathrm{C}$)} 
  & RMSE & 0.1868 $\pm$ 0.0114 & 0.1723 $\pm$ 0.0093 & 0.1682 $\pm$ 0.0074 & 0.5080 $\pm$ 0.0124 & 0.5025 $\pm$ 0.0202 & \textbf{0.1243 $\pm$ 0.0045} \\
& & MAE  & 0.1398 $\pm$ 0.0073 & 0.1203 $\pm$ 0.0051 & 0.1270 $\pm$ 0.0031 & 0.3978 $\pm$ 0.0080 & 0.3986 $\pm$ 0.0089 & \textbf{0.0941 $\pm$ 0.0022} \\
\cmidrule(lr){2-9}
& \multirow{2}{*}{Outlet Temperature ($^\circ\mathrm{C}$)} 
  & RMSE & 0.1976 $\pm$ 0.0051 & 0.2692 $\pm$ 0.0186 & 0.3301 $\pm$ 0.0281 & 0.6480 $\pm$ 0.0009 & 0.6478 $\pm$ 0.0111 & \textbf{0.1852 $\pm$ 0.0169} \\
& & MAE  & 0.1457 $\pm$ 0.0076 & 0.2122 $\pm$ 0.0173 & 0.2476 $\pm$ 0.0228 & 0.5228 $\pm$ 0.0008 & 0.5213 $\pm$ 0.0130 & \textbf{0.1382 $\pm$ 0.0161} \\
\midrule
\multirow{6}{*}{SM2} 
& \multirow{2}{*}{Flow Rate ($\mathrm{l/min}$)} 
  & RMSE & 0.1211 $\pm$ 0.0029 & \textbf{0.1152 $\pm$ 0.0021} & 0.1871 $\pm$ 0.0003 & 0.1160 $\pm$ 0.0004 & 0.1169 $\pm$ 0.0036 & 0.1247 $\pm$ 0.0027 \\
& & MAE  & \textbf{0.0728 $\pm$ 0.0039} & 0.0777 $\pm$ 0.0030 & 0.1288 $\pm$ 0.0009 & 0.0733 $\pm$ 0.0009 & 0.0731 $\pm$ 0.0043 & 0.0821 $\pm$ 0.0032 \\
\cmidrule(lr){2-9}
& \multirow{2}{*}{Inlet Temperature ($^\circ\mathrm{C}$)} 
  & RMSE & 0.1347 $\pm$ 0.0039 & 0.1547 $\pm$ 0.0047 & 0.1302 $\pm$ 0.0132 & 0.2836 $\pm$ 0.0049 & 0.2836 $\pm$ 0.0095 & \textbf{0.1091 $\pm$ 0.0019} \\
& & MAE  & 0.1036 $\pm$ 0.0054 & 0.1123 $\pm$ 0.0053 & 0.1031 $\pm$ 0.0109 & 0.2207 $\pm$ 0.0046 & 0.2218 $\pm$ 0.0074 & \textbf{0.0853 $\pm$ 0.0017} \\
\cmidrule(lr){2-9}
& \multirow{2}{*}{Outlet Temperature ($^\circ\mathrm{C}$)} 
  & RMSE & 0.0786 $\pm$ 0.0012 & 0.0996 $\pm$ 0.0036 & 0.1014 $\pm$ 0.0209 & 0.2293 $\pm$ 0.0047 & 0.2358 $\pm$ 0.0115 & \textbf{0.0605 $\pm$ 0.0099} \\
& & MAE  & 0.0620 $\pm$ 0.0008 & 0.0771 $\pm$ 0.0031 & 0.0792 $\pm$ 0.0166 & 0.1781 $\pm$ 0.0043 & 0.1850 $\pm$ 0.0113 & \textbf{0.0472 $\pm$ 0.0087} \\
\bottomrule
\end{tabular}
}
\end{table*}

\begin{table*}[ht]
\centering
\caption{Dataset 2: RMSE and MAE comparison across methods (mean $\pm$ std over 3 runs)}
\label{tab:dataset2_opt}
\resizebox{\linewidth}{!}{%
\begin{tabular}{lllcccccc}
\toprule
\multirow{2}{*}{} & \multirow{2}{*}{Sensor} & \multirow{2}{*}{Metric} 
& \multicolumn{6}{c}{Methods} \\
\cmidrule(lr){4-9}
& & & GRU-GCN & LSTM & 1D-CNN & DGC & GCN & HSTGNN \\
\midrule
\multirow{6}{*}{SM1} 
& \multirow{2}{*}{Flow Rate ($\mathrm{l/min}$)} 
  & RMSE & \textbf{0.1561 $\pm$ 0.0030} & 0.1689 $\pm$ 0.0030 & 0.2487 $\pm$ 0.0014 & 0.1668 $\pm$ 0.0024 & 0.1646 $\pm$ 0.0020 & 0.1680 $\pm$ 0.0048 \\
& & MAE  & \textbf{0.0806 $\pm$ 0.0046} & 0.0952 $\pm$ 0.0018 & 0.1493 $\pm$ 0.0012 & 0.0910 $\pm$ 0.0045 & 0.0869 $\pm$ 0.0027 & 0.0856 $\pm$ 0.0035 \\
\cmidrule(lr){2-9}
& \multirow{2}{*}{Inlet Temperature ($^\circ\mathrm{C}$)} 
  & RMSE & 0.3384 $\pm$ 0.0059 & 0.2024 $\pm$ 0.0256 & 0.2261 $\pm$ 0.0169 & 0.5713 $\pm$ 0.0011 & 0.5746 $\pm$ 0.0008 & \textbf{0.2003 $\pm$ 0.0135} \\
& & MAE  & 0.2415 $\pm$ 0.0018 & 0.1512 $\pm$ 0.0201 & 0.1623 $\pm$ 0.0108 & 0.4463 $\pm$ 0.0071 & 0.4499 $\pm$ 0.0088 & \textbf{0.1471 $\pm$ 0.0132} \\
\cmidrule(lr){2-9}
& \multirow{2}{*}{Outlet Temperature ($^\circ\mathrm{C}$)} 
  & RMSE & 0.2351 $\pm$ 0.0125 & 0.2849 $\pm$ 0.0121 & 0.2848 $\pm$ 0.0053 & 0.5952 $\pm$ 0.0149 & 0.5957 $\pm$ 0.0050 & \textbf{0.1987 $\pm$ 0.0134} \\
& & MAE  & 0.1782 $\pm$ 0.0127 & 0.2029 $\pm$ 0.0105 & 0.2167 $\pm$ 0.0080 & 0.4089 $\pm$ 0.0179 & 0.4074 $\pm$ 0.0024 & \textbf{0.1556 $\pm$ 0.0065} \\
\midrule
\multirow{6}{*}{SM2} 
& \multirow{2}{*}{Flow Rate ($\mathrm{l/min}$)} 
  & RMSE & 0.0628 $\pm$ 0.0005 & 0.0992 $\pm$ 0.0006 & 0.1687 $\pm$ 0.0021 & 0.0626 $\pm$ 0.0009 & \textbf{0.0609 $\pm$ 0.0017} & 0.0821 $\pm$ 0.0029 \\
& & MAE  & 0.0472 $\pm$ 0.0009 & 0.0727 $\pm$ 0.0008 & 0.1149 $\pm$ 0.0021 & 0.0468 $\pm$ 0.0010 & \textbf{0.0455 $\pm$ 0.0012} & 0.0563 $\pm$ 0.0010 \\
\cmidrule(lr){2-9}
& \multirow{2}{*}{Inlet Temperature ($^\circ\mathrm{C}$)} 
  & RMSE & 0.1533 $\pm$ 0.0059 & 0.1704 $\pm$ 0.0044 & 0.1452 $\pm$ 0.0169 & 0.3140 $\pm$ 0.0011 & 0.3120 $\pm$ 0.0054 & \textbf{0.1204 $\pm$ 0.0069} \\
& & MAE  & 0.1148 $\pm$ 0.0069 & 0.1183 $\pm$ 0.0016 & 0.1018 $\pm$ 0.0037 & 0.2423 $\pm$ 0.0013 & 0.2415 $\pm$ 0.0052 & \textbf{0.0755 $\pm$ 0.0034} \\
\cmidrule(lr){2-9}
& \multirow{2}{*}{Outlet Temperature ($^\circ\mathrm{C}$)} 
  & RMSE & 0.0634 $\pm$ 0.0011 & 0.1330 $\pm$ 0.0142 & 0.0785 $\pm$ 0.0012 & 0.1965 $\pm$ 0.0032 & 0.1941 $\pm$ 0.0029 & \textbf{0.0609 $\pm$ 0.0021} \\
& & MAE  &  0.0491 $\pm$ 0.0004 & 0.0948 $\pm$ 0.0124 & 0.0585 $\pm$ 0.0013 & 0.1461 $\pm$ 0.0026 & 0.1447 $\pm$ 0.0028 & \textbf{0.0483 $\pm$ 0.0009} \\
\bottomrule
\end{tabular}
}
\end{table*}

\begin{table*}[ht]
\centering
\caption{Dataset 3: RMSE and MAE comparison across methods (mean $\pm$ std over 3 runs)}
\label{tab:dataset3_opt}
\resizebox{\linewidth}{!}{%
\begin{tabular}{lllcccccc}
\toprule
\multirow{2}{*}{} & \multirow{2}{*}{Sensor} & \multirow{2}{*}{Metric} 
& \multicolumn{6}{c}{Methods} \\
\cmidrule(lr){4-9}
& & & GRU-GCN & LSTM & 1D-CNN & DGC & GCN & HSTGNN \\
\midrule
\multirow{6}{*}{SM1} 
& \multirow{2}{*}{Flow Rate ($\mathrm{l/min}$)} 
  & RMSE & 0.1447 $\pm$ 0.0010 & 0.2244 $\pm$ 0.0215 & 0.2450 $\pm$ 0.0090 & 0.1586 $\pm$ 0.0018 & 0.1572 $\pm$ 0.0068 & \textbf{0.1439 $\pm$ 0.0031} \\
& & MAE  & \textbf{0.0762 $\pm$ 0.0005} & 0.1516 $\pm$ 0.0216 & 0.1641 $\pm$ 0.0112 & 0.0930 $\pm$ 0.0015 & 0.0903 $\pm$ 0.0095 & 0.0781 $\pm$ 0.0059 \\
\cmidrule(lr){2-9}
& \multirow{2}{*}{Inlet Temperature ($^\circ\mathrm{C}$)} 
  & RMSE & 0.2931 $\pm$ 0.0211 & 0.2465 $\pm$ 0.0112 & 0.2544 $\pm$ 0.0144 & 0.6284 $\pm$ 0.0258 & 0.6302 $\pm$ 0.0069 & \textbf{0.2247 $\pm$ 0.0132} \\
& & MAE  & 0.2240 $\pm$ 0.0165 & 0.1834 $\pm$ 0.0146 & 0.1852 $\pm$ 0.0096 & 0.4372 $\pm$ 0.0213 & 0.4391 $\pm$ 0.0011 & \textbf{0.1565 $\pm$ 0.0021} \\
\cmidrule(lr){2-9}
& \multirow{2}{*}{Outlet Temperature ($^\circ\mathrm{C}$)} 
  & RMSE & 0.2369 $\pm$ 0.0118 & 0.2983 $\pm$ 0.0132 & 0.2661 $\pm$ 0.0139 & 0.4738 $\pm$ 0.0107 & 0.4543 $\pm$ 0.0254 & \textbf{0.2320 $\pm$ 0.0136} \\
& & MAE  & 0.1574 $\pm$ 0.0147 & 0.2280 $\pm$ 0.0122 & 0.1897 $\pm$ 0.0150 & 0.3416 $\pm$ 0.0083 & 0.3253 $\pm$ 0.0165 & \textbf{0.1553 $\pm$ 0.0161} \\
\midrule
\multirow{6}{*}{SM2} 
& \multirow{2}{*}{Flow Rate ($\mathrm{l/min}$)} 
  & RMSE & 0.0998 $\pm$ 0.0012 & 0.1890 $\pm$ 0.0307 & 0.2049 $\pm$ 0.0173 & 0.1072 $\pm$ 0.0006 & 0.1069 $\pm$ 0.0007 & \textbf{0.0991 $\pm$ 0.0049} \\
& & MAE  & \textbf{0.0566 $\pm$ 0.0014} & 0.1375 $\pm$ 0.0213 & 0.1394 $\pm$ 0.0157 & 0.0679 $\pm$ 0.0008 & 0.0681 $\pm$ 0.0013 & 0.0571 $\pm$ 0.0041 \\
\cmidrule(lr){2-9}
& \multirow{2}{*}{Inlet Temperature ($^\circ\mathrm{C}$)} 
  & RMSE & 0.3033 $\pm$ 0.0132 & 0.2288 $\pm$ 0.0021 & 0.2384 $\pm$ 0.0042 & 0.3893 $\pm$ 0.0068 & 0.3942 $\pm$ 0.0122 & \textbf{0.1944 $\pm$ 0.0062} \\
& & MAE  & 0.2093 $\pm$ 0.0069 & 0.1702 $\pm$ 0.0021 & 0.1823 $\pm$ 0.0045 & 0.3087 $\pm$ 0.0055 & 0.3120 $\pm$ 0.0095 & \textbf{0.1330 $\pm$ 0.0009} \\
\cmidrule(lr){2-9}
& \multirow{2}{*}{Outlet Temperature ($^\circ\mathrm{C}$)} 
  & RMSE & 0.0997 $\pm$ 0.0096 & 0.2046 $\pm$ 0.0129 & 0.1244 $\pm$ 0.0082 & 0.2413 $\pm$ 0.0053 & 0.2638 $\pm$ 0.0116 & \textbf{0.0983 $\pm$ 0.0102} \\
& & MAE  & 0.0743 $\pm$ 0.0066 & 0.1531 $\pm$ 0.0114 & 0.0981 $\pm$ 0.0070 & 0.1844 $\pm$ 0.0066 & 0.2048 $\pm$ 0.0122 & \textbf{0.0734 $\pm$ 0.0095} \\
\bottomrule
\end{tabular}
}
\end{table*}

\begin{table*}[ht]
\centering
\caption{Dataset 4: RMSE and MAE comparison across methods (mean $\pm$ std over 3 runs)}
\label{tab:dataset4_opt}
\resizebox{\linewidth}{!}{%
\begin{tabular}{lllcccccc}
\toprule
\multirow{2}{*}{} & \multirow{2}{*}{Sensor} & \multirow{2}{*}{Metric} 
& \multicolumn{6}{c}{Methods} \\
\cmidrule(lr){4-9}
& & & GRU-GCN & LSTM & 1D-CNN & DGC & GCN & HSTGNN \\
\midrule
\multirow{6}{*}{SM1} 
& \multirow{2}{*}{Flow Rate ($\mathrm{l/min}$)} 
  & RMSE & 0.1934 $\pm$ 0.0006 & 0.1908 $\pm$ 0.0016 & 0.2424 $\pm$ 0.0009 & 0.2137 $\pm$ 0.0012 & 0.2127 $\pm$ 0.0005 & \textbf{0.1803 $\pm$ 0.0054} \\
& & MAE  & 0.1018 $\pm$ 0.0002 & 0.1208 $\pm$ 0.0050 & 0.1492 $\pm$ 0.0013 & 0.1192 $\pm$ 0.0024 & 0.1187 $\pm$ 0.0004 & \textbf{0.0974 $\pm$ 0.0010} \\
\cmidrule(lr){2-9}
& \multirow{2}{*}{Inlet Temperature ($^\circ\mathrm{C}$)} 
  & RMSE & 0.2355 $\pm$ 0.0006 & 0.2478 $\pm$ 0.0020 & 0.2365 $\pm$ 0.0086 & 0.5202 $\pm$ 0.0235 & 0.5119 $\pm$ 0.0093 & \textbf{0.1722 $\pm$ 0.0126} \\
& & MAE  & 0.1838 $\pm$ 0.0004 & 0.1976 $\pm$ 0.0035 & 0.1884 $\pm$ 0.0060 & 0.3986 $\pm$ 0.0251 & 0.3909 $\pm$ 0.0080 & \textbf{0.1292 $\pm$ 0.0102} \\
\cmidrule(lr){2-9}
& \multirow{2}{*}{Outlet Temperature ($^\circ\mathrm{C}$)} 
  & RMSE & 0.1846 $\pm$ 0.0029 & 0.2519 $\pm$ 0.0233 & 0.2407 $\pm$ 0.0298 & 0.4392 $\pm$ 0.0044 & 0.4400 $\pm$ 0.0044 & \textbf{0.1656 $\pm$ 0.0170} \\
& & MAE  & 0.1343 $\pm$ 0.0018 & 0.1866 $\pm$ 0.0288 & 0.1851 $\pm$ 0.0234 & 0.3296 $\pm$ 0.0028 & 0.3320 $\pm$ 0.0030 & \textbf{0.1157 $\pm$ 0.0047} \\
\midrule
\multirow{6}{*}{SM2} 
& \multirow{2}{*}{Flow Rate ($\mathrm{l/min}$)} 
  & RMSE & 0.1730 $\pm$ 0.0006 & 0.1660 $\pm$ 0.0016 & 0.2286 $\pm$ 0.0033 & 0.1879 $\pm$ 0.0020 & 0.1866 $\pm$ 0.0006 & \textbf{0.1590 $\pm$ 0.0048} \\
& & MAE  & 0.0951 $\pm$ 0.0038 & 0.1002 $\pm$ 0.0022 & 0.1421 $\pm$ 0.0023 & 0.1067 $\pm$ 0.0016 & 0.1064 $\pm$ 0.0015 & \textbf{0.0874 $\pm$ 0.0054} \\
\cmidrule(lr){2-9}
& \multirow{2}{*}{Inlet Temperature ($^\circ\mathrm{C}$)} 
  & RMSE & 0.1225 $\pm$ 0.0045 & 0.1672 $\pm$ 0.0031 & 0.1670 $\pm$ 0.0042 & 0.3311 $\pm$ 0.0108 & 0.3273 $\pm$ 0.0112 & \textbf{0.1115 $\pm$ 0.0018} \\
& & MAE  & 0.0921 $\pm$ 0.0018 & 0.1252 $\pm$ 0.0049 & 0.1272 $\pm$ 0.0038 & 0.2681 $\pm$ 0.0106 & 0.2652 $\pm$ 0.0113 & \textbf{0.0808 $\pm$ 0.0026} \\
\cmidrule(lr){2-9}
& \multirow{2}{*}{Outlet Temperature ($^\circ\mathrm{C}$)} 
  & RMSE & 0.0848 $\pm$ 0.0013 & 0.1225 $\pm$ 0.0009 & 0.1060 $\pm$ 0.0019 & 0.2024 $\pm$ 0.0019 & 0.2064 $\pm$ 0.0028 & \textbf{0.0740 $\pm$ 0.0039} \\
& & MAE  & 0.0634 $\pm$ 0.0011 & 0.0957 $\pm$ 0.0021 & 0.0843 $\pm$ 0.0016 & 0.1524 $\pm$ 0.0018 & 0.1566 $\pm$ 0.0020 & \textbf{0.0551 $\pm$ 0.0024} \\
\bottomrule
\end{tabular}
}
\end{table*}

\subsection{Qualitative Results}

To complement the quantitative evaluation, we provide a qualitative comparison of predicted and measured sensor signals across representative datasets and smart meters. The figures visualize how different models track temporal dynamics, offering insight into their ability to capture both smooth trends and rapid fluctuations. This qualitative analysis highlights differences in stability, responsiveness, and error patterns that are not fully reflected by aggregate error metrics alone.

\begin{figure*} \centering \includegraphics[width=\linewidth]{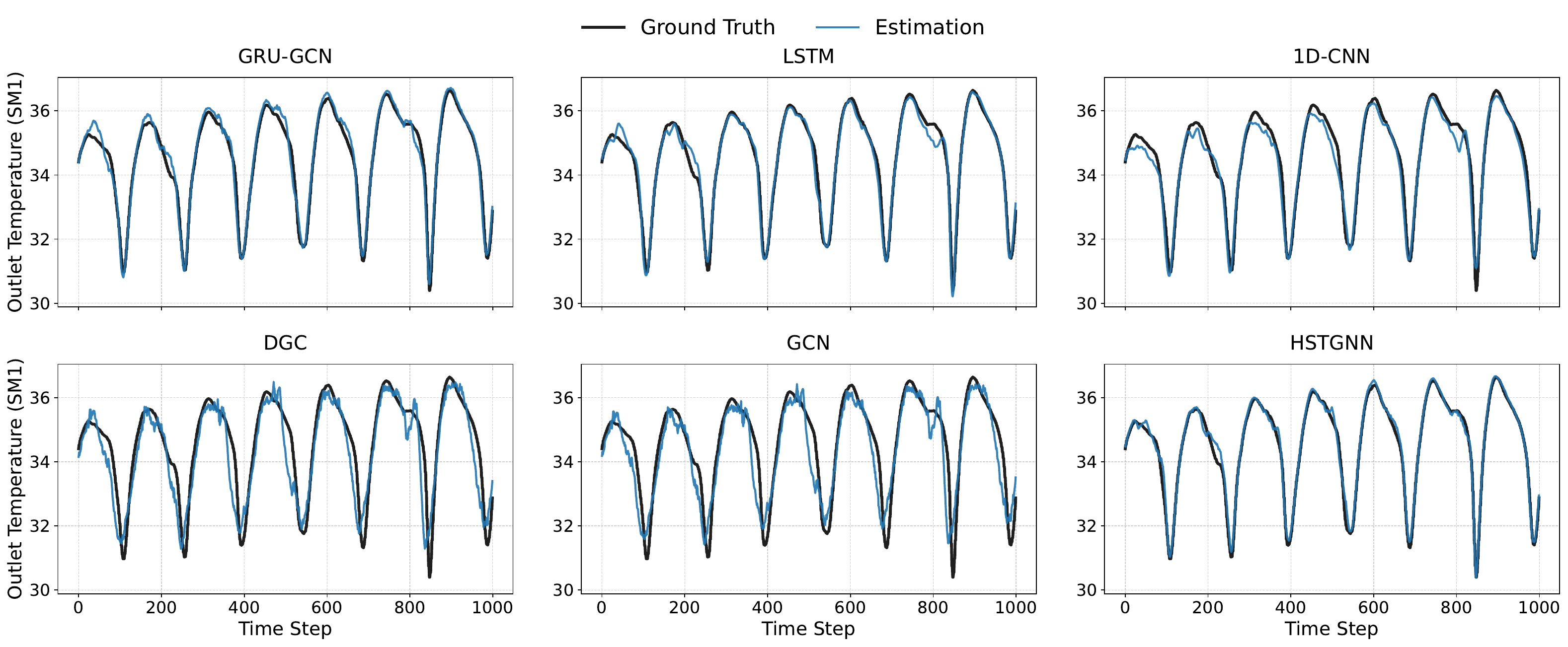} \caption{Outlet temperature prediction for SM~1, Dataset~1. HSTGNN closely follows the true signal; GRU-GCN is second best; graph-only models show larger deviations.} \label{DS01Vis} 

\end{figure*}

Figure~\ref{DS01Vis} illustrates the prediction results for the outlet temperature of Smart Meter~1 in Dataset~1. The HSTGNN model closely follows the ground-truth temperature trajectory and exhibits the smallest deviation over time. GRU-GCN provides the second-best performance, capturing the overall trend but with slightly larger discrepancies during periods around local peaks. In contrast, purely graph-based methods such as DGC and GCN show noticeable deviations from the true signal, particularly around local peaks, indicating a reduced ability to capture temporal dynamics.

\begin{figure*} \centering \includegraphics[width=\linewidth]{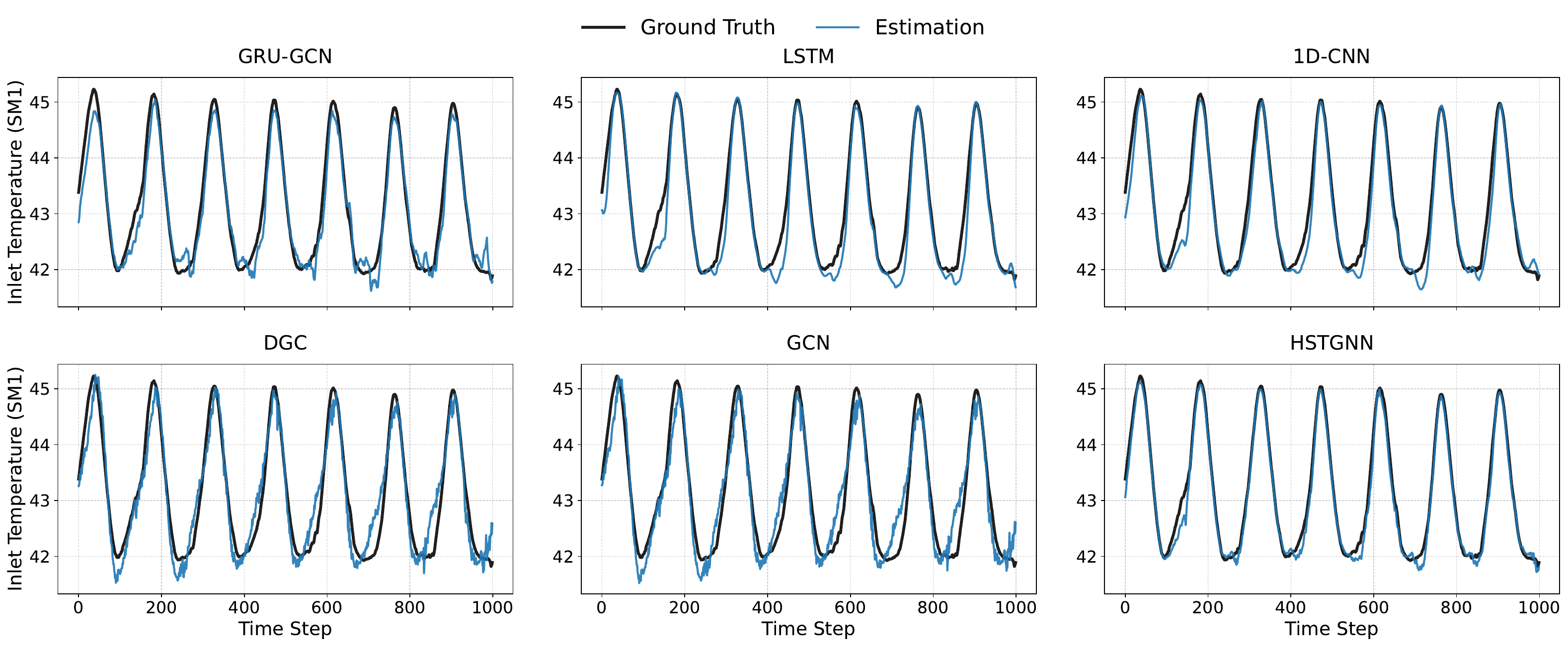} \caption{Inlet temperature for SM~1, Dataset~2. HSTGNN is most accurate; temporal-only and graph-only models deviate during transients.} \label{DS02Vis} 

\end{figure*}

Figure~\ref{DS02Vis} presents the inlet temperature predictions for Smart Meter~1 in Dataset~2. The overall behavior is consistent with that observed in Dataset~1, where HSTGNN produces the most accurate and stable estimates. GRU-GCN again captures the dominant temporal patterns, but exhibits increased smoothing compared to HSTGNN. Temporal-only and graph-only baselines display larger errors, particularly during transient temperature changes, highlighting the benefit of joint spatial-temporal modeling.

\begin{figure*} \centering \includegraphics[width=\linewidth]{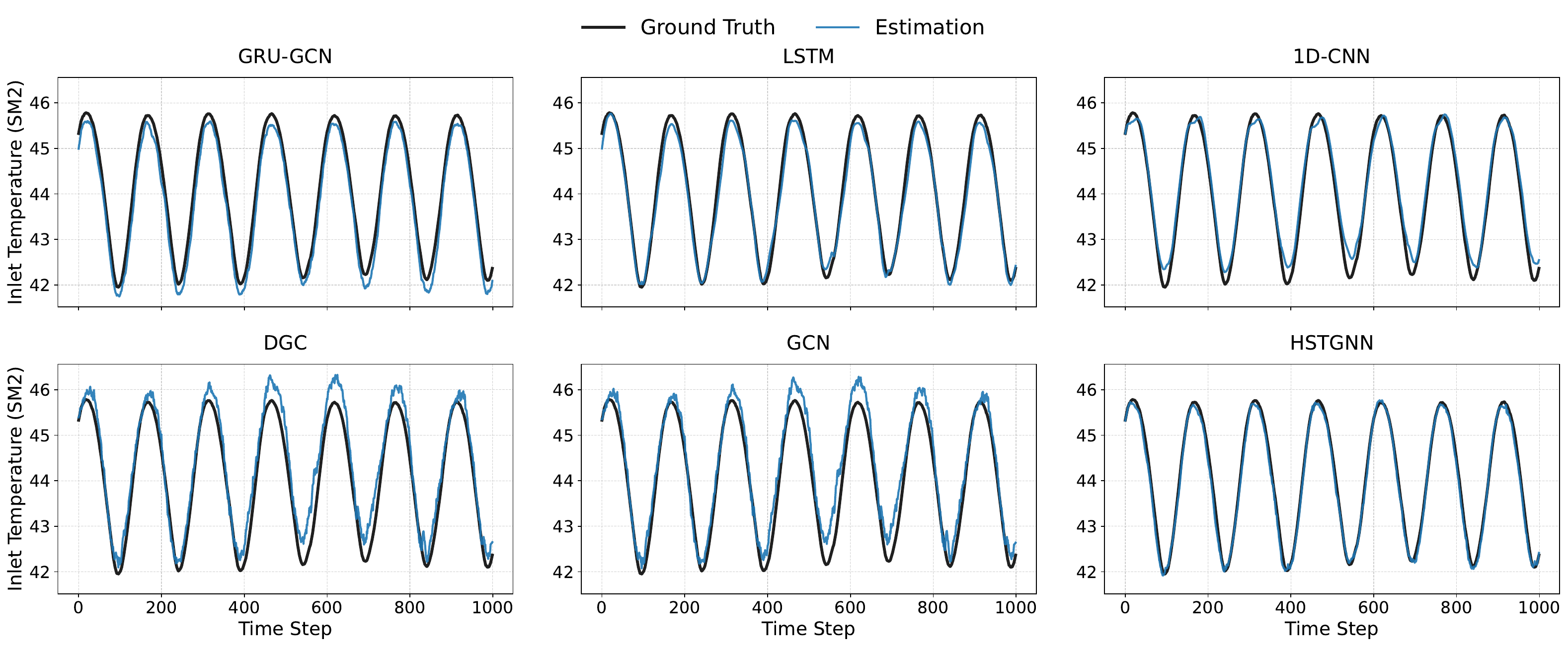} \caption{Inlet temperature for SM~2, Dataset~3. HSTGNN tracks the measured signal well; GRU-GCN captures general trends.} \label{DS03Vis} 

\end{figure*}

Figure~\ref{DS03Vis} shows the inlet temperature predictions for Smart Meter~2 in Dataset~3. As in the previous cases, HSTGNN aligns closely with the measured signal across the entire time horizon. GRU-GCN remains competitive but demonstrates reduced accuracy during high-variability intervals. The graph-based baselines fail to consistently track peaks, reinforcing the importance of integrating spatial context with temporal modeling.

\begin{figure*} \centering \includegraphics[width=\linewidth]{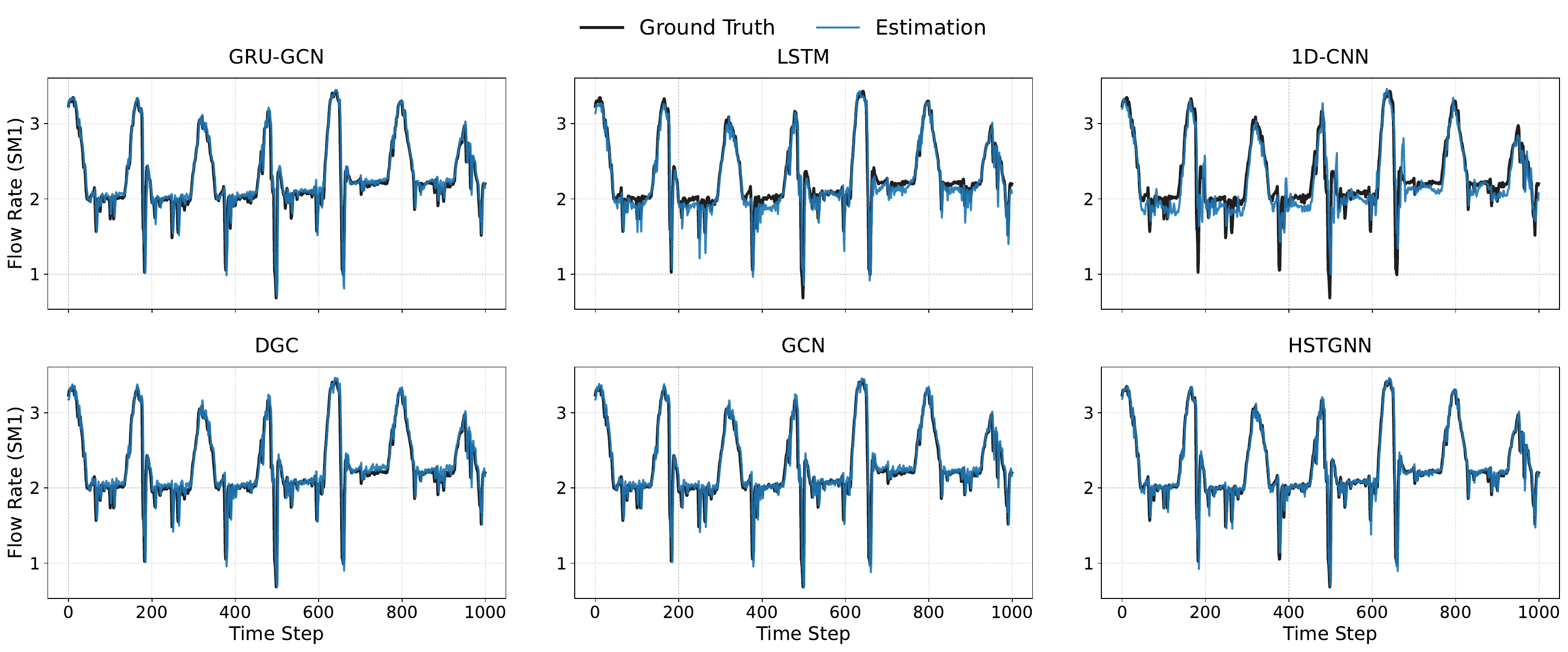} \caption{Flow rate for SM~1, Dataset~4. HSTGNN and GRU-GCN perform best; graph-only models outperform temporal-only baselines.} \label{DS04Vis} 

\end{figure*}

Figure~\ref{DS04Vis} shows the flow rate signal for Smart Meter~1 in Dataset~4. Both HSTGNN and GRU-GCN accurately capture the magnitude and temporal evolution of the flow signal, with HSTGNN exhibiting slightly improved stability. Graph-based methods perform better than purely temporal baselines in this case, suggesting that flow dynamics benefit from spatial information to capture rapid fluctuations. Nevertheless, the strongest performance is achieved when spatial and temporal dependencies are modeled jointly.

\subsection{Ablation Study}

To evaluate the contribution of each sensor type, an ablation study is conducted by systematically removing one sensor group at a time, namely flow, pressure, and temperature measurements. The performance of the resulting models is compared against the full HSTGNN model, as summarized in Table \ref{tab:ablation}. The results show that removing temperature measurements leads to the most significant degradation in performance, particularly for temperature prediction tasks. For instance, the RMSE for inlet temperature at SM1 increases from 0.1243 to 0.7013, and similar trends are observed across all temperature-related variables. This highlights the critical role of temperature information in capturing the underlying thermo-hydraulic behavior of the network.

\begin{table}
\centering
\caption{Ablation study of HSTGNN by removing sensor groups (Dataset 1)}
\label{tab:ablation}
\resizebox{0.8\linewidth}{!}{%
\begin{tabular}{lllcccc}
\toprule
\multirow{2}{*}{} & \multirow{2}{*}{Sensor} & \multirow{2}{*}{Metric} 
& \multicolumn{4}{c}{Ablation Variants} \\
\cmidrule(lr){4-7}
& & & Full Model & w/o Pressure & w/o Flow & w/o Temperature \\
\midrule
\multirow{6}{*}{SM1} 
& \multirow{2}{*}{Flow Rate ($\mathrm{l/min}$)} 
  & RMSE & \textbf{0.1856} & 0.1863 & 0.3439 & 0.1859 \\
& & MAE  & \textbf{0.1017} & 0.1021 & 0.2162 & 0.1090 \\
\cmidrule(lr){2-7}
& \multirow{2}{*}{Inlet Temperature ($^\circ\mathrm{C}$)} 
  & RMSE & \textbf{0.1243} & 0.1598 & 0.2031 & 0.7013 \\
& & MAE  & \textbf{0.0941} & 0.1247 & 0.1522 & 0.5139 \\
\cmidrule(lr){2-7}
& \multirow{2}{*}{Outlet Temperature ($^\circ\mathrm{C}$)} 
  & RMSE & \textbf{0.1852} & 0.2040 & 0.2760 & 0.7307 \\
& & MAE  & \textbf{0.1382} & 0.1551 & 0.2128 & 0.5373 \\
\midrule
\multirow{6}{*}{SM2} 
& \multirow{2}{*}{Flow Rate ($\mathrm{l/min}$)} 
  & RMSE & \textbf{0.1247} & 0.1291 & 0.1375 & 0.1279 \\
& & MAE  & \textbf{0.0821} & 0.0864 & 0.0919 & 0.0855 \\
\cmidrule(lr){2-7}
& \multirow{2}{*}{Inlet Temperature ($^\circ\mathrm{C}$)} 
  & RMSE & \textbf{0.1091} & 0.1490 & 0.1169 & 0.8251 \\
& & MAE  & \textbf{0.0853} & 0.1145 & 0.0893 & 0.6014 \\
\cmidrule(lr){2-7}
& \multirow{2}{*}{Outlet Temperature ($^\circ\mathrm{C}$)} 
  & RMSE & \textbf{0.0605} & 0.1000 & 0.1008 & 0.4195 \\
& & MAE  & \textbf{0.0472} & 0.0832 & 0.0855 & 0.3302 \\
\bottomrule
\end{tabular}
}
\end{table}

The ablation results also provide insight into the physical role of each sensor type. Removing temperature measurements causes the largest degradation because the virtual smart meter targets include inlet and outlet temperatures, and these variables directly describe the thermal state of the network. Without temperature inputs, the model must infer thermal behavior indirectly from hydraulic variables, which is insufficient during boiler on-off cycles and transient heat propagation. This explains the substantial RMSE increase for temperature-related targets, for example from 0.1243 to 0.7013 for SM1 inlet temperature and from 0.1091 to 0.8251 for SM2 inlet temperature.

Excluding flow measurements also degrades performance, especially for flow prediction at SM1, where the RMSE increases from 0.1856 to 0.3439. This is expected because flow is the main carrier of thermal energy in the network and determines the advection of temperature changes from the heat station toward consumer substations. Therefore, flow measurements are not only important for predicting flow targets, but also help explain temperature dynamics through the thermo-hydraulic coupling $\dot{q}=\dot{m}c_p(T_s-T_r)$. The deterioration of several temperature predictions after removing flow sensors confirms that the model uses hydraulic information to support thermal inference.

In contrast, removing pressure measurements leads to a comparatively smaller performance drop under the present laboratory conditions. This does not imply that pressure is unimportant in general. Rather, the experimental setup is relatively compact and pressure-controlled, so much of the relevant hydraulic state may already be encoded by the available flow and temperature measurements. In larger networks with longer pipe sections, multiple pumps, control valves, and stronger hydraulic interactions, pressure measurements are expected to play a more important role. Overall, the ablation study confirms that the multi-branch design is physically meaningful: temperature sensors dominate thermal-state reconstruction, flow sensors capture advective transport and load variation, and pressure sensors provide complementary hydraulic context.

While these results confirm the necessity of a multi-branch modeling, the effectiveness of the model also depends on how these distinct physical properties are modeled. To this end, Table~\ref{tab:dataset4_ablation} presents a secondary ablation study evaluating the impact of HSTGNN’s key architectural components. Specifically, we compare the full model against a minimal variant that replaces heterogeneous branching, sensor-specific spatial modeling, and node identity embeddings with a generic linear encoder and self-attention mechanism. By dropping these explicit structural and temporal inductive biases, we can quantify the value of specialized graph structures. The results, reported in terms of RMSE and MAE, show that the full HSTGNN consistently outperforms the simplified variant across all sensors, validating the importance of architectural specialization when modeling complex spatial-temporal dynamics.

\begin{table}[ht]
\centering
\caption{Ablation study on Dataset 4: HSTGNN compared to a minimal variant without heterogeneous branching, sensor-type spatial modeling, or sensor-specific embeddings.}
\label{tab:dataset4_ablation}
\resizebox{0.5\linewidth}{!}{%
\begin{tabular}{lllcc}
\toprule
\multirow{2}{*}{} & \multirow{2}{*}{Sensor} & \multirow{2}{*}{Metric} 
& \multicolumn{2}{c}{Models} \\
\cmidrule(lr){4-5}
& & & HSTGNN & Simplified \\
\midrule
\multirow{6}{*}{SM1} 
& \multirow{2}{*}{Flow Rate ($\mathrm{l/min}$)} 
  & RMSE & \textbf{0.1803} & 0.2124 \\
& & MAE  & \textbf{0.0974} & 0.1259 \\
\cmidrule(lr){2-5}
& \multirow{2}{*}{Inlet Temperature ($^\circ\mathrm{C}$)} 
  & RMSE & \textbf{0.1722} & 0.3734 \\
& & MAE  & \textbf{0.1292} & 0.2809 \\
\cmidrule(lr){2-5}
& \multirow{2}{*}{Outlet Temperature ($^\circ\mathrm{C}$)} 
  & RMSE & \textbf{0.1656} & 0.3658 \\
& & MAE  & \textbf{0.1157} & 0.2797 \\
\midrule
\multirow{6}{*}{SM2} 
& \multirow{2}{*}{Flow Rate ($\mathrm{l/min}$)} 
  & RMSE & \textbf{0.1590} & 0.1788 \\
& & MAE  & \textbf{0.0874} & 0.1007 \\
\cmidrule(lr){2-5}
& \multirow{2}{*}{Inlet Temperature ($^\circ\mathrm{C}$)} 
  & RMSE & \textbf{0.1115} & 0.2236 \\
& & MAE  & \textbf{0.0808} & 0.1710 \\
\cmidrule(lr){2-5}
& \multirow{2}{*}{Outlet Temperature ($^\circ\mathrm{C}$)} 
  & RMSE & \textbf{0.0740} & 0.2092 \\
& & MAE  & \textbf{0.0551} & 0.1631 \\
\bottomrule
\end{tabular}
}
\end{table}

\subsection{Interpretability of Learned Graph Structures and Cross-Type Fusion}
\label{sec:learned_graph_interpretability}

District heating networks are governed by coupled thermo-hydraulic processes. Temperature dynamics are affected by heat-station operation, transport delay, heat losses, return-flow mixing, and the advective effect of mass flow. Pressure and flow dynamics are governed by pump operation, valve states, branch resistance, and consumer-side hydraulic conditions. Therefore, a physically meaningful data-driven model should not treat temperature, pressure, and flow measurements as independent signal groups. To examine whether the proposed HSTGNN learns physically plausible internal structures, we analyze both the learned sensor-type-specific graphs and the cross-type attention layer.

It is important to clarify the interpretation of these learned structures. The learned graphs are not intended to reconstruct the physical pipe topology directly. Instead, each learned edge represents a data-driven functional dependency that is useful for predicting the target smart-meter variables. A high-scoring edge may correspond to a direct physical relationship, but it may also reflect an indirect dependency caused by shared flow paths, boiler control, pump operation, recirculation, return-flow effects, or correlated operating regimes. Similarly, the cross-type attention weights are computed after type-specific temporal and graph encoding. Therefore, this analysis should be viewed as a qualitative interpretability check rather than a validation of the underlying thermo-hydraulic governing equations.

\begin{figure*}[ht]
    \centering
    \includegraphics[width=\linewidth]{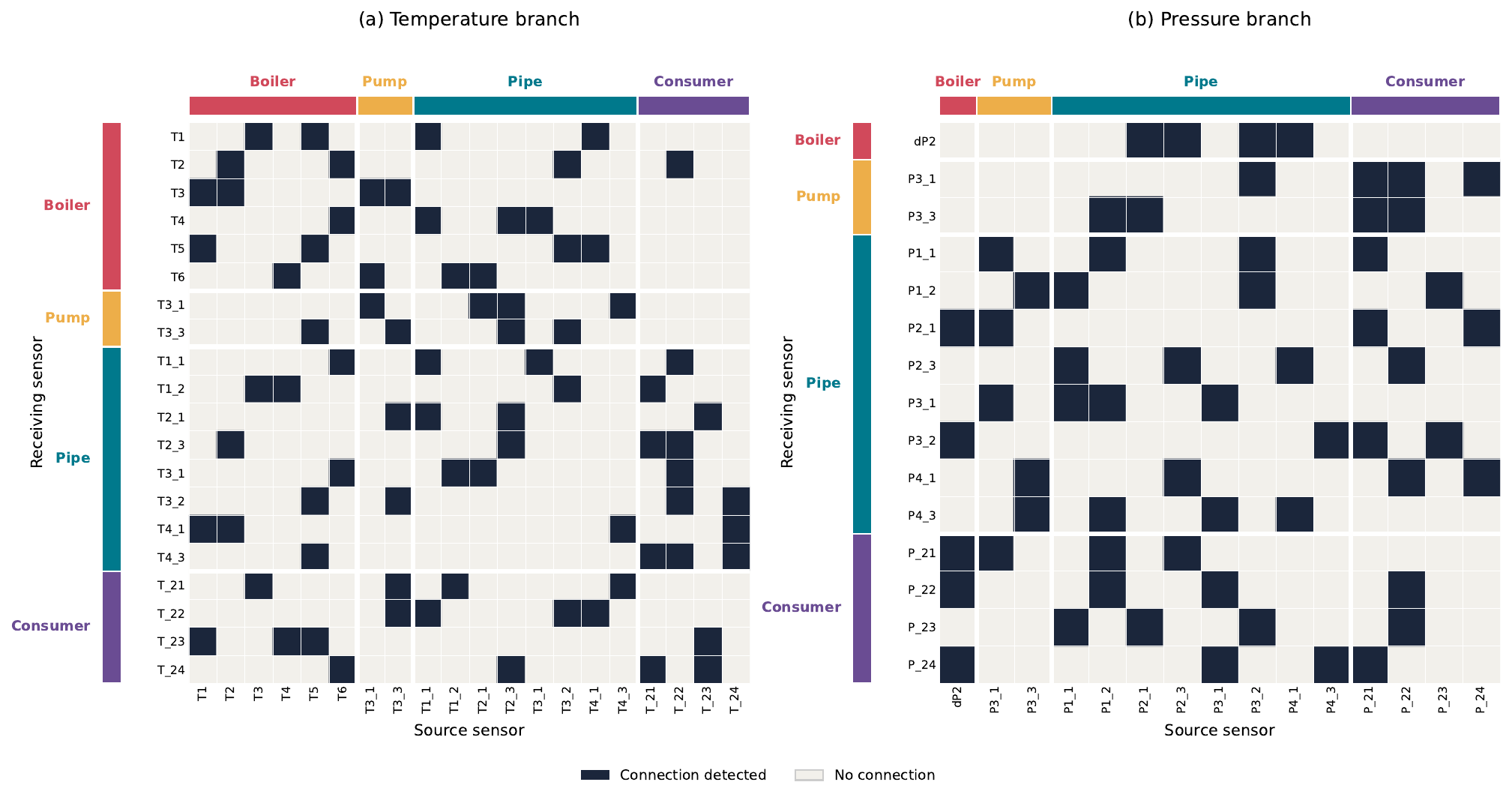}
    \caption{Learned top-$K$ functional adjacency matrices for the temperature and pressure branches. Rows denote receiving sensors and columns denote source sensors. Non-zero entries correspond to top-$4$ dependencies selected from the learned score matrices. The learned graphs are interpreted as data-driven functional dependency structures rather than direct reconstructions of the pipe topology.}
    \label{fig:learned_graphs}
\end{figure*}

Figure~\ref{fig:learned_graphs} shows the learned top-$K$ functional adjacency matrices for the temperature and pressure branches, where $K=4$ incoming edges are retained for each receiving sensor for visualization. Rows denote receiving sensors and columns denote source sensors. The learned temperature graph contains several dependencies that are consistent with the SWIL topology and operating behavior. For example, learned edges from pipe-unit temperature sensors to consumer-side sensors, such as $\mathrm{Pipe\_T2\_3}\rightarrow\mathrm{Consumer\_T\_24}$, $\mathrm{Pipe\_T1\_2}\rightarrow\mathrm{Consumer\_T\_21}$, and $\mathrm{Pipe\_T3\_2}\rightarrow\mathrm{Consumer\_T\_22}$, are consistent with thermal transport between the pipe unit and consumer-side measurements. The graph also identifies heat-station and pipe-unit dependencies, such as $\mathrm{Boiler\_T4}\rightarrow\mathrm{Pipe\_T1\_2}$, and within heat-station dependencies, such as $\mathrm{Boiler\_T1}\rightarrow\mathrm{Boiler\_T3}$ and $\mathrm{Boiler\_T5}\rightarrow\mathrm{Boiler\_T1}$. These relationships are broadly consistent with the expected thermal behavior of the testbed, where temperature changes propagate through the experimental loop with transport delay and are influenced by supply/return mixing, boiler operation, and consumer-side load changes.

At the same time, some learned temperature edges connect consumer-side sensors back to pipe-unit, pumping-loop, or boiler-side sensors. These edges should not be interpreted as direct downstream thermal propagation. Instead, they likely reflect indirect functional dependencies caused by return-flow effects, load variation, boiler on--off control, pump operation, and correlated operating regimes. This behavior is expected in a closed-loop district heating testbed, where the thermal state at one location can provide information about the operating state of other parts of the loop even when the sensors are not direct physical neighbors.

The learned pressure graph exhibits a more distributed structure than the temperature graph. This is physically reasonable because pressure disturbances propagate much faster than temperature changes and are coupled through the hydraulic state of the entire network. Several high-scoring pressure dependencies involve the heat-station differential pressure sensor, pumping-unit pressure sensors, pipe-unit pressure sensors, and consumer-side pressure sensors. For example, the learned pressure graph includes dependencies such as $\mathrm{Pipe\_P3\_1}\rightarrow\mathrm{Consumer\_P\_24}$, $\mathrm{Pump\_P3\_1}\rightarrow\mathrm{Pipe\_P1\_1}$, $\mathrm{Pipe\_P1\_1}\rightarrow\mathrm{Consumer\_P\_23}$, $\mathrm{Boiler\_dP2}\rightarrow\mathrm{Consumer\_P\_21}$, and $\mathrm{Pump\_P3\_1}\rightarrow\mathrm{Pipe\_P3\_1}$. These dependencies are consistent with hydraulic coupling among the heat station, pump, pipe-unit branches, and consumer-side hydraulic states. Unlike temperature propagation, pressure dependencies are not expected to follow a simple downstream path; they are naturally more global and bidirectional because pressure states are jointly affected by pump operation, branch resistance, valve states, and consumer-side hydraulic conditions.

To complement the graph interpretation, we also inspect the global cross-type attention weights after the branch-specific graph encoders. For each query--key sensor-type pair, the attention values are first averaged over all sensor pairs in the corresponding block, and then each query row is normalized to sum to one. This aggregation reduces the influence of unequal group sizes and emphasizes the relative average attention strength assigned to each modality during fusion.

The resulting global attention matrix is shown in Table~\ref{tab:global_attention_blockmean_norm}. Temperature queries assign the largest relative attention to flow keys. This does not mean that temperature measurements are less important than flow measurements, because temperature information has already been processed by the temperature-specific temporal and graph branch before cross-type fusion. Rather, it suggests that encoded flow representations provide important complementary information for refining temperature representations, which is physically plausible because mass flow governs advective heat transport in district heating systems. Pressure queries assign the largest relative attention to pressure keys, consistent with the hydraulic nature of pressure dynamics. Flow queries assign the largest relative attention to flow keys, indicating that flow-specific information remains dominant during the fusion stage. These patterns are consistent with physically plausible information routing in the fusion layer, while still preserving modality-specific pressure and flow information.

\begin{table}[ht]
\centering
\caption{Global row-normalized type-level block-mean attention. Rows denote query sensor types and columns denote key sensor types. Values are computed after type-specific temporal and graph encoding and are interpreted as information-routing diagnostics rather than causal physical dependencies.}
\label{tab:global_attention_blockmean_norm}
\begin{tabular}{lccc}
\toprule
\textbf{Query / Key} & \textbf{Temperature} & \textbf{Pressure} & \textbf{Flow} \\
\midrule
Temperature & 0.326 & 0.225 & \textbf{0.449} \\
Pressure    & 0.270 & \textbf{0.454} & 0.276 \\
Flow        & 0.306 & 0.299 & \textbf{0.395} \\
\bottomrule
\end{tabular}
\end{table}

Overall, the learned graph analysis provides the main physical interpretability evidence. The temperature graph captures functional dependencies related to heat-station operation, pipe-unit transport, consumer-side thermal variation, pumping-loop interaction, and return-flow effects. The pressure graph captures distributed hydraulic coupling among heat-station, pump, pipe, and consumer pressure sensors. The attention analysis provides an additional diagnostic showing that the fusion layer combines information from all three sensor modalities in a physically plausible manner. However, neither the learned graphs nor the attention weights are interpreted as direct physical topology, causal dependency matrices, or recovered governing equations. A deeper transient-state thermo-hydraulic analysis would require controlled excitation experiments or labeled events that isolate pump, valve, load, and thermal transport effects, which is beyond the scope of the present study.

\subsection{Potential Real-World Applications}

The proposed HSTGNN framework can support several practical applications in operational district heating systems. First, it can provide continuous virtual heat metering at branches or consumer substations where physical smart meters are unavailable, temporarily disconnected, or sampled at low frequency. This can improve network observability for energy accounting, load monitoring, and operational supervision without requiring dense meter deployment at every location. However, in applications involving regulatory-grade billing, the proposed virtual sensing framework is intended to complement, rather than replace, certified physical heat meters.

Second, the model can support measurement quality monitoring and metrological quality assurance. By comparing physical meter readings with virtual estimates generated from the surrounding SCADA sensor pool, large or persistent residuals may indicate sensor drift, calibration degradation, communication failures, hardware faults, or inconsistent meter behaviour. Such residual-based monitoring can help utilities identify meters that require inspection, recalibration, or maintenance.

Third, HSTGNN can contribute to real-time operational monitoring and anomaly detection. Deviations between estimated and measured system states may reveal abnormal flow conditions, unexpected thermal losses, hydraulic disturbances, or changes in consumer-side operation. Therefore, the model can provide an additional data-driven layer for supervising the thermo-hydraulic state of the network.

Finally, HSTGNN can be integrated into district heating digital twin platforms. In this role, virtual measurements can fill observability gaps and provide fast state estimates for online monitoring, what-if analysis, pump scheduling, return-temperature management, and predictive operational control. The proposed framework does not replace physics-based simulation; rather, it complements it by providing fast data-driven estimates from available heterogeneous SCADA measurements.

\section{Conclusion}
\label{sec:conclusion}

This work presented a sensor-type-aware spatial-temporal graph neural network for virtual sensing in district heating systems. The main methodological contribution is a heterogeneous modeling framework in which flow, temperature, and pressure measurements are processed through separate spatial-temporal branches. This design allows the model to learn type-specific spatial structures and temporal dynamics before combining them through an attention-based fusion mechanism for joint reconstruction of unobserved sensor readings.  The proposed framework addresses the challenges posed by sparse instrumentation and the complex coupling of physical variables by explicitly accounting for heterogeneous sensor behavior in district heating networks. The extensive evaluation results demonstrate that the proposed method consistently outperforms a range of baseline models. These results confirm that explicit modeling of heterogeneous sensor types improves both prediction accuracy and robustness. Another important contribution of this study is the development of a controlled laboratory dataset collected at the Aalborg Smart Water Infrastructure Laboratory. The dataset provides high-resolution, synchronized measurements of flow, temperature, and pressure under representative district heating operating conditions. Since such experimental datasets are rarely available in the literature, where research often relies on simulations or incomplete field measurements, this resource fills a relevant gap and offers a reproducible benchmark for future studies.

At the same time, the proposed architecture has some limitations. In the current work, each branch follows the same processing structure for all sensor types, which provides a consistent design, but may not fully exploit sensor-specific modeling requirements. Furthermore, in real deployments, communication interruptions, calibration drift, battery constraints, or hardware faults may lead to missing or unreliable input measurements. Such failures could degrade the accuracy of the learned spatial-temporal dependencies, especially if multiple sensors of the same type fail simultaneously. Future work will therefore evaluate robustness under random sensor masking, block missingness, and complete sensor-group outages. Potential solutions include training with sensor dropout, adding missingness masks to the input representation, incorporating imputation modules inspired by missing-variable time-series models, and using uncertainty estimates to flag unreliable virtual meter predictions. Moreover, a current limitation of the field is the lack of publicly available benchmark datasets for virtual sensing in district heating networks. To the best of our knowledge, no public dataset currently provides synchronized pressure, flow, temperature, and consumer smart-meter measurements required to develop and fairly compare virtual sensing methods. Addressing this gap is one of the primary contributions of this work. The proposed laboratory testbed was therefore specifically designed to provide high-resolution, synchronized multi-physics measurements together with reliable ground truth under controlled operating conditions, enabling reproducible evaluation and future benchmarking. While this controlled environment is well suited for method development and systematic analysis, validation on larger, branched municipal district heating networks remains an important next step before real-world deployment. From a modeling perspective, HSTGNN is designed to scale to larger networks because temporal encoders and graph-learning modules share parameters across sensors of the same type, while the sparse top-$K$  neighbor selection restricts message passing to the most relevant functional connections rather than requiring dense all-to-all communication. Consequently, scaling to larger systems primarily increases the number of node embeddings and sparse message-passing operations, whereas the number of learnable parameters remains largely unchanged. Future work will evaluate the proposed framework on larger operational district heating networks and high-fidelity digital-twins comprising additional consumer branches,  pumping stations, and more complex hydraulic topologies.

\section*{Data Availability}

The dataset supporting the findings of this study is publicly available on Zenodo: \\
\href{https://doi.org/10.5281/zenodo.23127439}
{https://doi.org/10.5281/zenodo.23127439}.

\section*{Acknowledgement}
This work was funded by the Swiss Federal Institute of Metrology (METAS). The authors thank Kamstrup for providing the experimental hardware.

 \bibliographystyle{elsarticle-num} 
 \bibliography{main}

\end{document}